\documentclass[lettersize,journal]{IEEEtran}
\usepackage{amsmath,amsfonts}
\usepackage{algorithmic}
\usepackage{algorithm}
\usepackage{array}
\usepackage[caption=false,font=normalsize,labelfont=sf,textfont=sf]{subfig}
\usepackage{textcomp}
\usepackage{stfloats}
\usepackage{url}
\usepackage{verbatim}
\usepackage{graphicx}
\usepackage{cite}

\usepackage{enumitem}
\usepackage{xcolor}
\usepackage{booktabs}
\usepackage{multirow}
\usepackage{mathrsfs}

\usepackage{caption}
\DeclareCaptionLabelFormat{tablelabel}
{\MakeUppercase{#1} \MakeUppercase{#2}}
\begin{document}

% \title{A Sample Article Using IEEEtran.cls\\ for IEEE Journals and Transactions}
\title{DE$^3$-BERT: Distance-Enhanced Early Exiting for BERT \\Based on Prototypical Networks}

\author{Jianing He, Qi Zhang, Weiping Ding,~\IEEEmembership{Senior Member,~IEEE}, Duoqian Miao, Jun Zhao, Liang Hu, \\and Longbing Cao,~\IEEEmembership{Senior Member,~IEEE}
        % <-this % stops a space
\thanks{This work has been submitted to the IEEE for possible publication. Copyright may be transferred without notice, after which this version may no longer be accessible. This work was supported in part by the National Natural Science Foundation of China under Grant 62376198; and in part by the Shanghai Baiyulan Pujiang Project under Grant 08002360429. \textit{(Corresponding author: Duoqian Miao.)}}% <-this % stops a space
\thanks{Jianing He is with the School of Information Engineering, Tianjin University of Commerce, Tianjin 300134, China (email: hejianing@tjcu.edu.cn).}
\thanks{Qi Zhang, Duoqian Miao, and Liang Hu are with the School of Computer Science and Technology, Tongji University, Shanghai 201804, China (email: zhangqi\_cs@tongji.edu.cn; dqmiao@tongji.edu.cn; lianghu@tongji.edu.cn).}
\thanks{Weiping Ding is with the School of Artificial Intelligence and Computer Science, Nantong University, Nantong 226019, China, and also the Faculty of Data Science, City University of Macau, Macau 999078, China (email: dwp9988@163.com).}
\thanks{Jun Zhao is with the National Laboratory of Pattern Recognition, Institute of Automation, Chinese Academy of Sciences, Beijing 100190, China (email: jzhao@nlpr.ia.ac.cn).}
\thanks{Longbing Cao is with the DataX Research Centre, School of Computing, Macquarie University, Sydney, NSW 2109, Australia (email: LongBing.Cao@mq.edu.au).}
}

% The paper headers
\markboth{Journal of \LaTeX\ Class Files,~Vol.~14, No.~8, August~2021}%
{Shell \MakeLowercase{\textit{et al.}}: A Sample Article Using IEEEtran.cls for IEEE Journals}

% \IEEEpubid{0000--0000/00\$00.00~\copyright~2021 IEEE}
% Remember, if you use this you must call \IEEEpubidadjcol in the second
% column for its text to clear the IEEEpubid mark.

\maketitle

\begin{abstract}
Early exiting has demonstrated its effectiveness in accelerating the inference of pre-trained language models like BERT by dynamically adjusting the number of layers executed. However, most existing early exiting methods only consider local information from individual test samples to determine their exiting indicators, failing to leverage the global information offered by sample population. This leads to a suboptimal estimation of prediction correctness, yielding erroneous exiting decisions. To bridge the gap, we explore the necessity of effectively combining local and global information to ensure reliable early exiting during inference. Purposefully, we leverage prototypical networks to learn class prototypes and devise a distance metric between samples and class prototypes. This enables us to utilize global information for estimating the correctness of early predictions. On this basis, we propose a novel Distance-Enhanced Early Exiting framework for BERT (DE$^3$-BERT). DE$^3$-BERT implements a hybrid exiting strategy that supplements classic entropy-based local information with distance-based global information to enhance the estimation of prediction correctness for more reliable exiting decisions. Extensive experiments on the GLUE benchmark demonstrate that DE$^3$-BERT consistently outperforms state-of-the-art models under different speed-up ratios with minimal storage or computational overhead, yielding a better trade-off between model performance and inference efficiency. Further analysis confirms the effectiveness of each component in our framework, as well as its generalization capability.
\end{abstract}


\section{Introduction}
\IEEEPARstart{I}{n} recent years, large-scale pre-trained language models (PLMs), such as GPT~\cite{radford2019language}, BERT~\cite{bert,Shadi2024FakeND}, RoBERTa~\cite{abs-1907-11692}, and ALBERT~\cite{LanCGGSS20}, have brought significant improvements to natural language processing tasks. Despite the tremendous success, these transformer-based models still suffer from high memory and computational overhead in both training and inference. In particular, the prolonged inference latency hinders their deployment in edge devices or real-time scenarios. Besides, overthinking~\cite{Overthinking} poses another challenging issue for these large PLMs. It refers to the phenomenon where the model performs well in intermediate layers for easy samples but experiences a decline in performance in deeper layers, leading to redundant computation and performance degradation. 

To speed up the inference of PLMs, numerous attempts have been made by employing model compression to condense their sizes. These approaches encompass network pruning~\cite{ pruning4, pruning5}, weight quantization~\cite{quantification2, quantification3}, knowledge distillation~\cite{KD2,Yu2025ZeroCL,Ye2025CCDE}, and weight sharing~\cite{weightsharing1, weightsharing2}. These methods can effectively save memory usage and computational costs to accelerate inference. However, they often necessitate additional training costs to meet different acceleration requirements, leading to inflexible adjustments to the speed-up ratio.%in model compression compress a large PLM into a more compact version

\begin{figure}[!t]
\centering
\vspace{-3mm}
\includegraphics[width=2.5in]{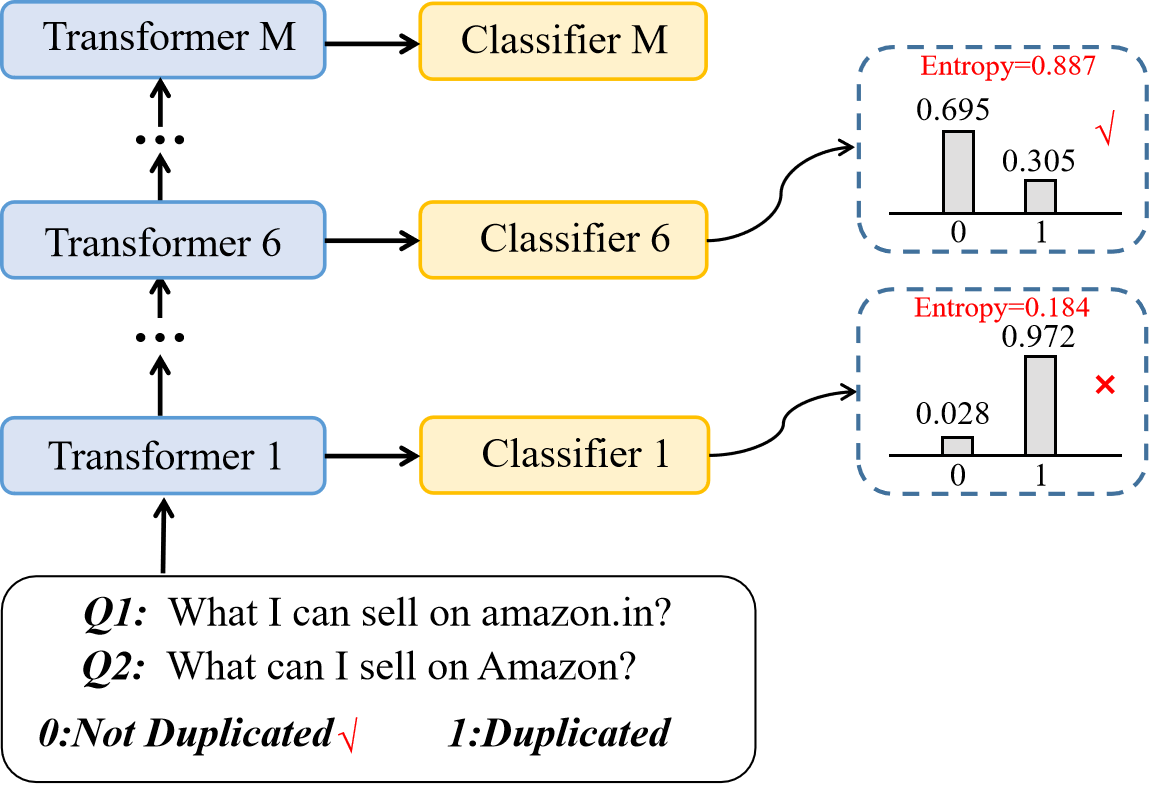}
\caption{The discrepancy between model confidence and prediction correctness. On a sample in QQP, the first internal classifier makes a wrong prediction with a low entropy value (high confidence), while the sixth classifier makes a correct prediction with a high entropy value (low confidence).}
\vspace{-3mm}
\label{fig:intro_entropy}
\end{figure}

To address these issues, early exiting~\cite{Deebert, Pabee, Berxit, PCEE, hashbased, NSP, twostageEE, li2021cascadebert, globalpast,ConsistentEE,COSEE} has been applied to accelerate the inference of PLMs. Unlike model compression, early exiting can easily adapt to different acceleration requirements by simply adjusting the threshold, without incurring additional training costs. In Fig.~\ref{fig:intro_entropy}, we illustrate early exiting based on BERT, where an internal classifier is added to each intermediate layer of BERT. This enables the early exiting of samples when the early predictions from these internal classifiers are deemed sufficiently correct, eliminating the need to traverse the entire model. This strategy employs adaptive inference to deal with easy samples with shallow layers of BERT and process difficult samples with deeper layers. This approach effectively mitigates the overthinking problem and accelerates model inference while maintaining high accuracy.%\subref{fig:overview_classic}

A typical implementation for early exiting is to devise an early exiting indicator that reflects the correctness of early predictions (i.e., prediction correctness) to establish the exiting criteria. According to the types of exiting indicators, there are mainly three early exiting strategies for the dynamic exiting of BERT models. The first strategy is confidence-based early exiting (e.g., DeeBERT~\cite{Deebert}, RightTool~\cite{Righttool}, and CascadeBERT~\cite{li2021cascadebert}). It utilizes the model confidence, i.e., entropy or label score of the prediction probability distribution, to estimate the correctness of early predictions for exit decision-making. The second strategy is patience-based early exiting (e.g., PABEE~\cite{Pabee} and LeeBERT~\cite{zhu2021leebert}), relying on the cross-layer consistency to determine when to exit. Early exiting in the third type learns to generate an early exiting indicator for making exiting decisions (e.g., BERxiT~\cite{Berxit} and ConsistentEE~\cite{ConsistentEE}). However, a reliable exiting strategy is to exit the sample when its early prediction is sufficiently correct. Unfortunately, the absence of ground-truth labels during inference poses extreme challenges to estimating prediction correctness, which compromises the reliability of exiting decisions.

\begin{figure}[!t]
\centering
\vspace{-5mm}
\includegraphics[width=2.5in]{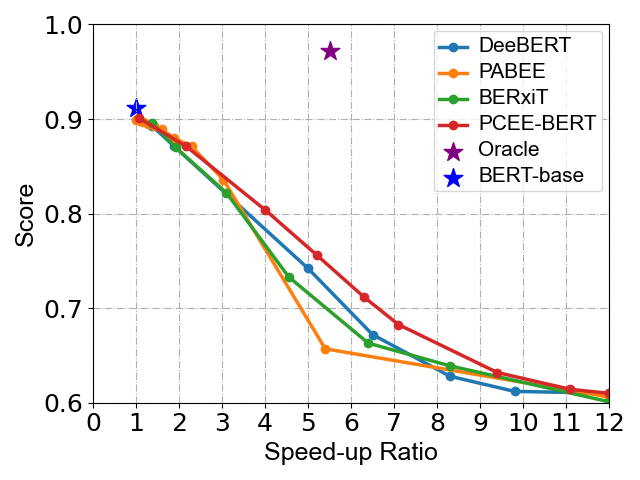}
\vspace{-1mm}
\caption{The performance-efficiency trade-offs of existing exiting strategies, the oracle, and the BERT-base backbone on the QNLI development set. Right and higher is better. The significant gap between the existing exiting strategies and the oracle indicates considerable potential for improving the estimation of prediction correctness, which is crucial to reliable exiting decisions.}
\vspace{-3mm}
\label{fig:oracle}
\end{figure}

In Fig.~\ref{fig:intro_entropy}, we illustrate the confidence-based exit decision-making process for a sample in the QQP task. Notably, we observe that a high-confidence prediction made by a classifier is not necessarily correct~\cite{Z22ood}. This indicates that model confidence may not always accurately reflect prediction correctness, potentially leading to erroneous exiting decisions. In Fig.~\ref{fig:oracle}, we further compare the existing exiting strategies with the original backbone model, as well as the oracle which is an ideal model that can always enable each sample to exit at the shallowest internal classifier that provides a correct label prediction~\cite{Righttool}. Essentially, the oracle serves as the upper bound for early exiting strategies as it provides the most accurate estimation of prediction correctness to achieve an optimal trade-off between model performance and efficiency. We can observe that the oracle outperforms the backbone model and existing exiting strategies by a large margin in both model performance and inference speed, which indicates significant room for improving the estimation of prediction correctness. We believe that the global information offered by the sample population can reflect the latent class information of test samples, which can be utilized to estimate the correctness of early predictions. However, the aforementioned strategies primarily focus on local information (e.g., entropy, consistency, and label scores) from individual test samples to formulate their early exiting indicators, while neglecting global information provided by the sample population. Hence, insufficient information exacerbates the difficulty of estimating prediction correctness, resulting in unreliable exiting decisions.

To bridge the gap, we explore the necessity of integrating both the local information from individual test samples and the global information provided by the sample population to accurately estimate prediction correctness and formulate reliable early exiting during inference. Inspired by prototype learning~\cite{snell2017prototypical} that involves learning class prototypes from a set of samples, we introduce a prototypical network paired with each internal classifier to learn class prototypes during training, as shown in Fig.~\ref{fig:overview}\subref{fig:overview_ours}. Then we devise a distance metric between samples and class prototypes to incorporate global information for estimating the correctness of early predictions. Accordingly, we propose a novel Distance-Enhanced Early Exiting framework for BERT (DE$^3$-BERT) based on prototypical networks. DE$^3$-BERT leverages distance-based global information to enhance the classic entropy-based early exiting through information fusion. Concretely, in DE$^3$-BERT, each prototypical network is implemented with a simple linear layer and is jointly trained with other components under an elaborate distance-aware regularization (DAR). During inference, we calculate distance scores between samples and class prototypes, and then integrate the distance-based global information with the classic entropy-based local information to enhance prediction correctness estimation for more reliable early exiting. To achieve this, a harmonic indicator that integrates global and local information is proposed to formulate a hybrid exiting strategy. Our contributions can be summarized as follows.

\begin{itemize}[leftmargin=*]  
\item{We disclose that the performance limitation of existing early exiting methods primarily stems from the discrepancy between exiting indicators and prediction correctness. We claim that this discrepancy arises from solely depending on local information from individual test samples to make exiting decisions while neglecting the global information provided by the sample population.}
\item{We leverage prototypical networks to learn class prototypes and devise a distance metric between samples and class prototypes to incorporate global information for estimating the prediction correctness from a global perspective.}
\item{We propose a novel Distance-Enhanced Early Exiting framework for BERT (DE$^3$-BERT) with a hybrid exiting strategy considering both the traditional entropy-based local information and the distance-based global information to facilitate the estimation of prediction correctness for making reliable exiting decisions.}
\item{Extensive experiments on the GLUE benchmark demonstrate that our DE$^3$-BERT achieves a superior performance-efficiency trade-off compared to state-of-the-art baseline methods, with minimal additional storage or computational overhead.
Further analysis validates the effectiveness of each component within our framework and confirms the necessity of incorporating global information to ensure accurate estimation of prediction correctness.
Additionally, our method shows insensitivity to parameter selection and demonstrates strong generality across out-of-distribution (OOD) data, different languages, and various backbone models.}
\end{itemize}
% We hope that this study will inspire enhancing the reliability of exiting decisions by exploiting the global information provided by the distance between samples.

The rest of this article is organized as follows. Section \ref{section:related_work} provides a review of relevant work. Section \ref{section:method} details our proposed DE$^3$-BERT framework. Extensive experimental results and comprehensive analyses are reported in Section \ref{section:experiments}. Finally, Section \ref{section:conclusion} concludes this article.

\section{RELATED WORK}
\label{section:related_work}
In this section, we review prior related works from three aspects, including static model compression, dynamic early exiting, and prototypical networks.%, which are the prior works of this article.

\subsection{Static Model Compression}
Model compression effectively compresses large PLMs to speed up inference while the compressed models remain static for all samples during inference. Specifically, network pruning~\cite{pruning4, pruning5} involves removing unnecessary weights and connections from the network. Weight quantization~\cite{quantification2, quantification3} reduces the precision of weight representations to save the computational and storage overhead of the model. Knowledge distillation~\cite{KD2,Yu2025ZeroCL,Ye2025CCDE} transfers knowledge from a larger teacher model to a smaller student model. Weight sharing~\cite{weightsharing1, weightsharing2} achieves model compression by sharing weights across different network components. However, compared with early exiting, these methods are less flexible to meet varying acceleration conditions, requiring additional training costs for speed-up ratio adjustments.%Various approaches have been proposed, including network pruning, weight quantization, knowledge distillation, and weight sharing. 

\subsection{Dynamic Early Exiting}
Early exiting, a parallel line of research for accelerating the inference of PLMs, is to dynamically stop inference for various input samples. DeeBERT~\cite{Deebert}, FastBERT~\cite{Fastbert}, and RightTool~\cite{Righttool} are pioneers in applying early exiting to accelerate the inference of BERT-style models. These studies employ the confidence level, such as entropy or label score of the prediction probability distribution, to estimate the correctness of early predictions for exit decision-making. On this basis, Lin \textit{et al.}~\cite{EEforSequenceLabeling} extended confidence-based early exiting from classification tasks to sequence labeling tasks. CeeBERT~\cite{ceebert} introduces an online learning algorithm to tackle threshold selection for confidence-based early exiting in cross-domain scenarios. To further improve the acceleration performance of early exiting models, one approach is to explore new exiting strategies for the inference stage. PABEE~\cite{Pabee} uses the cross-layer consistency to determine early exiting. PCEE-BERT~\cite{PCEE} employs a hybrid exiting strategy that combines confidence-based and patience-based strategies. BERxiT~\cite{Berxit} and ConsistentEE~\cite{ConsistentEE} leverage neural networks to generate exiting indicators for making exiting decisions. HASHEE~\cite{hashbased} utilizes hash functions to allocate each token to a designated exiting layer, providing a novel static paradigm for early exiting. In addition, an alternative approach to improve acceleration performance is to strengthen the capability of internal classifiers through various techniques, such as model calibration, architecture refinement, and optimized training schemes. CascadeBERT~\cite{li2021cascadebert} employs early exiting within multiple cascaded complete models, thus strengthening the representation capability of internal classifiers. Moreover, it incorporates a difficulty-aware objective to calibrate the prediction probability distribution. LeeBERT~\cite{zhu2021leebert} introduces a cross-layer distillation loss with learned weights to promote the training of internal classifiers. GPFEE~\cite{globalpast} integrates both past and future states as inputs for each internal classifier, facilitating high-quality early predictions.
ConsistentEE~\cite{ConsistentEE} minimizes the classification loss for each sample solely at its exiting layer, thus enhancing consistency between training and testing to improve the capability of each classifier.
COSEE~\cite{COSEE} assigns sample-wise loss weights for each classifier to bridge the gap between training and testing while enabling flexible adjustments to the speed-up ratio.

\subsection{Prototypical Networks}
Prototypical networks are first proposed in~\cite{snell2017prototypical}, which enable the learning of class prototypes from a set of training samples to classify the test sample based on the distances between the test sample and these prototypes. Prototypical networks provide an effective framework to learn a metric space for classification, which have been widely applied in few-shot image classification~\cite{imageclassification1,imageclassification2} and image recognition tasks~\cite{imagerecognition}. In recent years, prototypical networks have been further extended to the field of natural language processing, demonstrating successful applications in few-shot relation classification~\cite{relationclassification1,relationclassification2} and named entity recognition tasks~\cite{ner}. Additionally, prototypical networks have proven to be effective in the field of speech processing, particularly for speech recognition tasks in low-resource scenarios~\cite{speechrecognition}. Besides the few-shot learning domain, prototypical networks have also been successfully applied to domain adaptation~\cite{PrototypeDomainAdaption} and anomaly detection~\cite{PrototypeAnomalyDetection}. %, especially in the domain of few-shot learning

Despite significant efforts dedicated to early exiting methods, there is still ample potential for improvement. The existing early exiting approaches rely solely on local information from individual test samples to determine when to exit, overlooking the valuable global information conveyed by the sample population. This limited perspective can impact the accuracy of prediction correctness estimation, leading to unreliable exiting decisions. In contrast, inspired by prototype learning, we propose to leverage prototypical networks to enhance the classic entropy-based early exiting by incorporating global information based on a distance metric between the test sample and class prototypes. Our framework comprehensively considers both entropy-based local information and distance-based global information to improve the estimation of prediction correctness, thereby achieving a better trade-off between model performance and efficiency.%enabling more reliable early exiting and

\begin{figure*}[!t]
\centering
\vspace{-3mm}
\subfloat[Classic early exiting framework]{\includegraphics[width=2.5in]{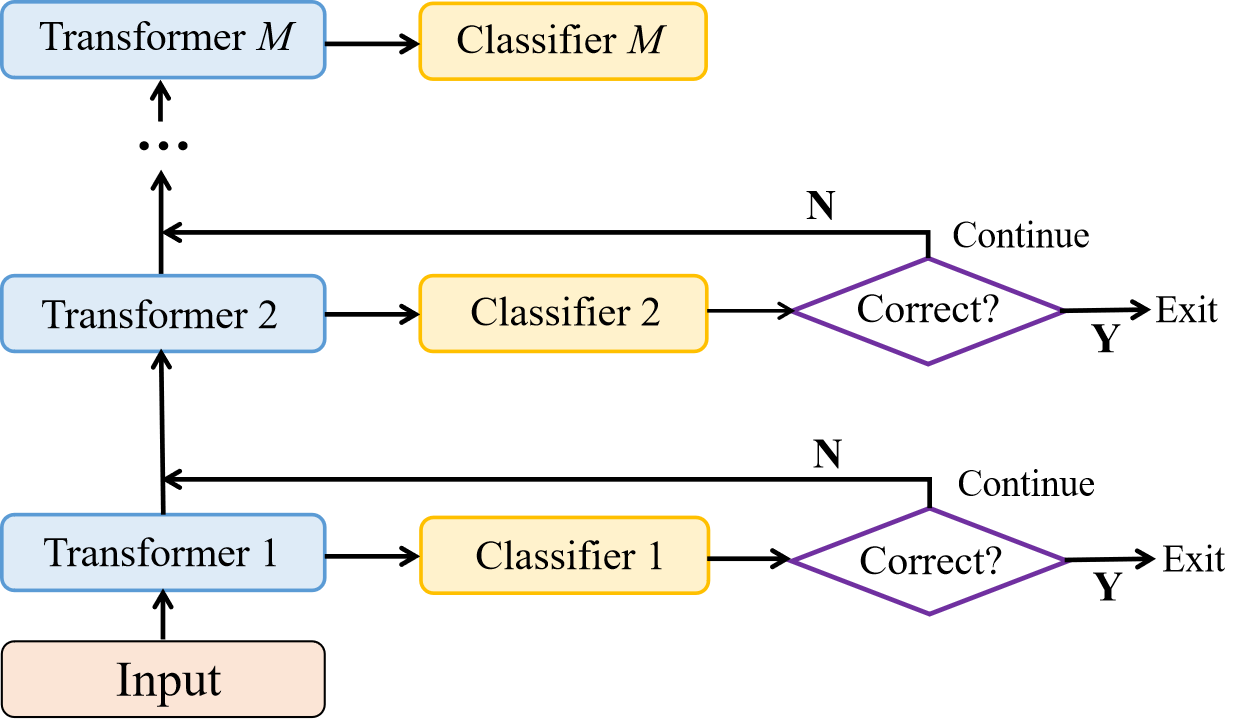}%
\label{fig:overview_classic}}
\hfil
\subfloat[Our proposed DE$^3$-BERT]{\includegraphics[height=1.5in]{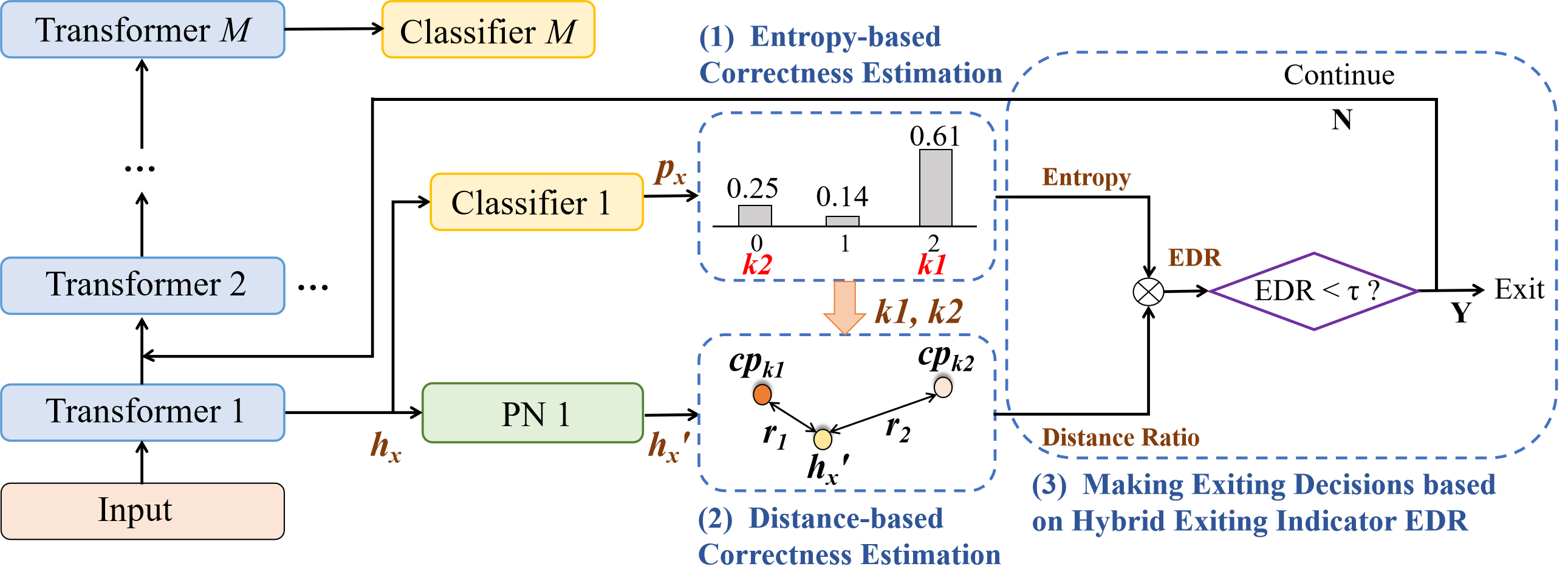}%
\label{fig:overview_ours}}
\caption{Comparison between the classic early exiting framework and our DE$^3$-BERT. Most existing frameworks rely on local information from individual test samples to estimate the correctness of early predictions. Instead, our DE$^3$-BERT considers both local information (entropy) and global information (distance ratio) to enhance the estimation of prediction correctness for more reliable exiting decisions. For the exit decision-making process of a sample $x$ shown in (b), (1) compute the entropy of the early prediction $p_x$ suggested by the internal classifier; (2) compute the distance ratio between the sample representation $h'_{x}$ and the two class prototypes ${cp}_{k1}$ and ${cp}_{k2}$ as illustrated in Equation~(\ref{dr}), where ${cp}_{k1}$ and ${cp}_{k2}$ correspond to the classes $k1$ and $k2$ with the highest two probability scores in $p_x$, and the prototypical network $\rm PN$ is used to map the hidden states $h_x$ into the metric space; (3) output the hybrid exiting indicator $\mathrm{EDR}$ according to Equation~(\ref{edr}), and terminate the inference process once the $\mathrm{EDR}$ value falls below the threshold $\tau$.}
\vspace{-3mm}
\label{fig:overview}
\end{figure*}

\section{The DE$^3$-BERT Framework}
\label{section:method}
\subsection{Overview}
\label{section:overview}

\subsubsection{Problem Definition}
Given a BERT-style model with $M$ layers, we denote the hidden states at the $m$th layer as $h^{(m)}$. To enable early exiting for the model's inference on a classification task with $K$ classes, each of its intermediate layers is attached with an internal classifier $f_m$ where $m\in\{1, 2, \cdots, M-1\}$ to give an early prediction $p^{(m)}=f_m(h^{(m)})$, i.e., a probability distribution over the $K$ classes.

\subsubsection{Framework Overview} 
We propose a novel Distance-Enhanced Early Exiting framework for BERT (DE$^3$-BERT) based on the prototypical networks, which leverages the distance-based global information to enhance the classic entropy-based early exiting, aiming for reliable exiting decisions with sufficiently correct early predictions. Fig.~\ref{fig:overview}\subref{fig:overview_ours} provides an overview of our framework. Structurally, we first add one prototypical network (paired with one internal classifier) into each intermediate layer of BERT to learn class prototype representations. Each prototypical network consists of a simple linear layer that maps the hidden states of each layer to a metric space for distance calculation. We further propose distance-aware regularization (DAR) as the loss function to jointly train the prototypical networks with the BERT. Then, we devise a normalized distance ratio to provide global information for estimating the prediction correctness by calculating the distances between the sample representation and class prototypes. Finally, we propose a hybrid exiting strategy with a harmonic indicator that integrates both local information (entropy) and global information (distance ratio) to improve the estimation of prediction correctness.% for more reliable early exiting.

\subsection{Training Scheme}
\label{section:training_scheme}

\subsubsection{Update of Class Prototypes}
Each intermediate layer owns a set of class prototypes, where each is updated using its corresponding class centroid through sliding average during training. Class centroids are obtained by averaging the sample representations class-by-class from each prototypical network separately. Accordingly, given the $m$th layer with $m\in\{1,2, \cdots, M-1\}$, the update strategy for its $k$th class prototype representation $k\in\{1,2,\cdots, K\}$ is formulated:
\begin{multline}
\label{cp-update}
cp_k^{(m)}[t]=(1-\gamma)\times cp_k^{(m)}[t-1]+\gamma\times c_k^{(m)}[t]
\end{multline}
where $\gamma$ is to adjust the strength of the update, $t$ denotes the update step, $cp_k^{(m)}$ and $c_k^{(m)}$ denote the class prototype and class centroid for the $k$th class at the $m$th layer, respectively.

\subsubsection{Distance-Aware Regularization}
We introduce distance-aware regularization (DAR) to train the prototypical networks and learn class prototype representations. Incorporated with the Center Loss \cite{centerloss,Hu2022DualDC} that minimizes the intra-class distances, DAR is formulated as the average distance between each sample representation and the prototype of its class:%with the Center Loss
\begin{multline}    
\label{DAR}
L_{\text{DAR}}^{(m)}=\frac{1}{N}\sum_{n=1}^N\left\|{h'}_n^{(m)},cp_{y_n}^{(m)}\right\|,
m\in\{1,\cdots, M-1\}
\end{multline}
where $N$ is the number of samples, and ${h'}_n^{(m)}$ denotes the representation of the $n$th sample derived from the $m$th prototypical network. $y_n$ is the ground-truth label of the $n$th sample, and $cp_{y_n}^{(m)}$ represents the class prototype corresponding to class $y_n$ at the $m$th layer. The metric $\|\cdot,\cdot\|$ is used to compute the cosine distance between ${h'}_n^{(m)}$ and $cp_{y_n}^{(m)}$:
\begin{equation}
\label{CosineDistance}
\left\|{h'}_n^{(m)},cp_{y_n}^{(m)}\right\|=1-\frac{{h'}_n^{(m)}\cdot cp_{y_n}^{(m)}}{\left\|{h'}_n^{(m)}\right\|\left\|cp_{y_n}^{(m)}\right\|}
\end{equation}
where $\cdot$ denotes the inner product, and $\|\cdot\|$ denotes the $\ell_2$ norm.
Note that the class prototypes are fixed when calculating DAR, i.e., no gradients back-propagating through $cp_{y_n}^{(m)}$.
Algorithm~\ref{algorithm:DAR_computation} outlines the computation of DAR for each intermediate layer.

In addition, we further consider two variants of DAR via incorporating with \textit{Alienation Loss} and \textit{Combined Loss}. The Alienation Loss is used to enlarge the difference between intra-class and inter-class distances. It is formulated as the average ratio between the distances from each sample to the prototype of its class and to the nearest prototype of a different class:
\begin{multline}\label{AlienationLoss}
L_{\text{DAR-A}}^{(m)}=\frac{1}{N}\sum_{n=1}^N\left[0.5\times \left(1+\frac{r_{n,y_n}^{(m)}-r_{n,z_n}^{(m)}}{\max\{r_{n,y_n}^{(m)},r_{n,z_n}^{(m)}\}}\right)\right]
\end{multline}
where $m\in\{1,2,\cdots, M-1\}$ denotes the number of layers, $N$ denotes the number of samples, $y_n$ denotes the ground-truth label for the $n$th sample, $r_{n,y_n}^{(m)}$ denotes the distance from the $n$th sample to its class prototype, and $r_{n,z_n}^{(m)}$ denotes the distance from the $n$th sample to the nearest prototype of a different class. The Combined Loss is utilized to reduce the intra-class distance and increase the difference between the intra-class and inter-class distances, which is formulated as a weighted sum of Center Loss and Alienation Loss:
\begin{equation}
\label{CombinedLoss}
L_{\text{DAR-CB}}^{(m)}=L_{\text{DAR-C}}^{(m)}+\beta\times L_{\text{DAR-A}}^{(m)}
\end{equation}
where $m\in\{1,2,\cdots, M-1\}$, $L_{\text{DAR-C}}^{(m)}$ denotes the Center Loss, $L_{\text{DAR-A}}^{(m)}$ denotes the Alienation Loss, and $\beta$ is a hyper-parameter used to balance the two losses. 

% For computational efficiency, we primarily use the Center Loss as DAR in this paper if not specified. We subsequently investigate the performance of different forms of DAR in Section \ref{section:DAR_comparison}.

\subsubsection{Training Objective} 
The cross-entropy loss is used to train the internal classifiers. Accordingly, the loss of each intermediate layer is formulated as the weighted sum of the cross-entropy loss and DAR:
\begin{equation}
\label{loss-m-1}
L^{(m)}=L_{\text{CE}}^{(m)}+\alpha\times L_{\text{DAR}}^{(m)}
\end{equation}
where $m\in\{1,2,\cdots, M-1\}$, $L_{\text{CE}}^{(m)}$ denotes the cross-entropy loss at the $m$th layer, and the regularization coefficient $\alpha$ is used to balance the cross-entropy loss and DAR. Since there is no need to exit early at the last layer, we minimize the cross-entropy loss for the classifier at the last ($M$th) layer:
\begin{equation}\label{loss-m}
L^{(M)}=L_{\text{CE}}^{(M)}.
\end{equation}

The total loss function is the weighted average of the losses across all layers:
\begin{equation}\label{total-loss}
L=\frac{\sum_{m=1}^M m\times L^{(m)}}{\sum_{m=1}^M m}
\end{equation}
where $L^{(m)}$ is the loss at the $m$th layer. Following PABEE~\cite{Pabee} and GPFEE~\cite{globalpast}, the weight of each layer is proportional to its layer number to balance parameter updates across shallow and deep layers. Neither prototypical networks nor internal classifiers share parameters across layers. The training process of DE$^3$-BERT is outlined in Algorithm~\ref{algorithm:training_process}.

\begin{algorithm}[!t]
	\caption{Computation of DAR for the $m$th layer.}
    \label{algorithm:DAR_computation}
	\begin{algorithmic}
		\STATE \textbf{Input:} Model, prototype representations for all $K$ classes $\{cp_k^{(m)}\}_{k=1}^K$, hidden states for all $N$ samples $\{h_n^{(m)}\}_{n=1}^N$, and their ground-truth labels $\{y_n\}_{n=1}^N$ 
		\STATE \textbf{Output:} DAR for the $m$th layer $L_{\text{DAR}}^{(m)}$
		\FOR{$n=1$ to $N$}
            \STATE \textcolor{green!50!black}{// Map the hidden states to the metric space}
            \STATE ${h'}_n^{(m)}=model.PrototypicalNetworks[m](h_n^{(m)})$;
            \STATE \textcolor{green!50!black}{// Extract the prototype representation of the true class}
            \STATE $cp_{y_n}^{(m)} \leftarrow cp_k^{(m)}|_{k=y_n}$;
            \STATE \textcolor{green!50!black}{// Compute the intra-class distance of the $n$th sample}
            \STATE $d_n^{(m)}=CosineDistance({h'}_n^{(m)},cp_{y_n}^{(m)})$;
		\ENDFOR
        \STATE \textcolor{green!50!black}{// Compute the average intra-class distance}
        \STATE $L_{\text{DAR}}^{(m)}=Average(\{d_n^{(m)}\}_{n=1}^N)$;
		\RETURN $L_{\text{DAR}}^{(m)}$;
	\end{algorithmic}
\end{algorithm}

\begin{algorithm}[!t]
	\caption{Training Process of DE$^3$-BERT.}
    \label{algorithm:training_process}
	\begin{algorithmic}
		\STATE \textbf{Input:} $M$-layer model, training dataset $D_{train}$, and total training steps $T$       
		\STATE \textbf{Output:} Model parameters $\Theta$ and class prototype representations for each intermediate layer $\{\{cp_k^{(m)}\}_{k=1}^K\}_{m=1}^{M-1}$
		\FOR{$t=1$ to $T$}
            \STATE Sample a mini-batch $B_t = \{(x_n, y_n)\}_{n=1}^{N_b}$ from $D_{train}$;
            \STATE \textcolor{green!50!black}{// Forward propagation}
            \FOR{$m = 1$ to $M-1$}
                \STATE \textcolor{green!50!black}{// Compute the cross-entropy loss}
                \STATE $L_{\text{CE}}^{(m)}=\frac{1}{N_b}\sum_{n=1}^{N_b}CrossEntropy(p_n^{(m)},y_n)$;
                \STATE \textcolor{green!50!black}{// Update the class prototype representations}
                \FOR{$k = 1$ to $K$}
                    \STATE $cp_k^{(m)} \leftarrow (1-\gamma)cp_k^{(m)}+\gamma\text{Mean}\big(\{{h'}_n^{(m)} \,|\, y_n = k\}\big)$;
                \ENDFOR
                \STATE \textcolor{green!50!black}{// Compute DAR}
                \STATE $L_{\text{DAR}}^{(m)}=\frac{1}{N_b}\sum_{n=1}^{N_b}CosineDistance({h'}_n^{(m)},cp_{y_n}^{(m)})$;
            \ENDFOR            
            \STATE \textcolor{green!50!black}{// Compute the cross-entropy loss for the $M$th layer}
            \STATE $L_{\text{CE}}^{(M)}=\frac{1}{N_b}\sum_{n=1}^{N_b}CrossEntropy(p_n^{(M)},y_n)$;
            \STATE \textcolor{green!50!black}{// Compute the total loss}
            \STATE $L=\frac{\sum_{m=1}^{M} mL_{\text{CE}}^{(m)}+\sum_{m=1}^{M-1} m\alpha L_{\text{DAR}}^{(m)}}{\sum_{m=1}^M m}$;
            \STATE \textcolor{green!50!black}{// Backward propagation and parameter update}
            \STATE $\Theta \leftarrow \text{AdamW}(\Theta, \nabla_{\Theta} L, \eta)$;
		\ENDFOR
		\RETURN $\Theta$, $\{\{cp_k^{(m)}\}_{k=1}^K\}_{m=1}^{M-1}$;
	\end{algorithmic}
\end{algorithm}

\subsection{Inference Strategy}
Building upon the classic entropy-based exiting framework, we incorporate the class prototypes to deliver reliable early exiting by jointly considering local and global information to estimate the correctness of early predictions.

\subsubsection{Entropy-based Correctness Estimation} 
The entropy provides local information that reflects the model's confidence level in its predictions for individual test samples. Following previous works~\cite{Deebert,globalpast,PCEE}, we use the classic normalized entropy to estimate the correctness of early predictions:

\begin{equation}
\label{normalized_entropy}
{\rm{Entropy}}(p^{(m)})=\frac{\sum_{k=1}^K p_k^{(m)}\times\log p_k^{(m)}}{\log (1/K)}
\end{equation}
where $p^{(m)}$ denotes the early prediction provided by the $m$th internal classifier, i.e. the probability distribution over the $K$ classes, $p_k^{(m)}$ denotes the probability score of class $k$ suggested by the $m$th internal classifier. The normalized entropy always falls between 0 and 1, and a lower entropy value indicates a higher confidence level of the internal classifier, typically leading to higher prediction correctness on the test sample.

\subsubsection{Distance-based Correctness Estimation} 
Apart from local information, we calculate the distances between the sample representation and class prototypes to provide global information for estimating the prediction correctness. Specifically, from the total $K$ classes, we select the classes with the highest two probability scores suggested by the $m$th internal classifier and then compute the distances between the test sample $x$ and the prototypes of these two classes respectively:
\begin{equation}
\label{r1r2}
r_1^{(m)}=\|h'^{(m)},cp_{k_1}^{(m)}\|,\,\,\,\,r_2^{(m)}=\|h'^{(m)},cp_{k_2}^{(m)}\|
\end{equation}
where $h'^{(m)}$ denotes the sample representation from the $m$th prototypical network. $cp_{k_1}^{(m)}$ and $cp_{k_2}^{(m)}$ denote the prototypes of the classes with the highest and the second highest probability scores at the $m$th layer, respectively, and $\|\cdot,\cdot\|$ denotes the cosine distance. Intuitively, if $r_1^{(m)}<r_2^{(m)}$, the distance metric result is consistent with the early prediction; otherwise, they are inconsistent. Therefore, we define a normalized distance ratio to describe the relative relationship between  $r_1^{(m)}$  and $r_2^{(m)}$, which effectively reflects the correctness of early predictions:
\begin{equation}\label{dr}
 {\rm{DR}}\left(r_1^{(m)}, r_2^{(m)}\right)=0.5\times \left(1+\frac{r_1^{(m)}-r_2^{(m)}}{\max\{r_1^{(m)},r_2^{(m)}\}}\right).
\end{equation}
The value of $\mathrm{DR}$ is between $0$ and $1$. A lower $\mathrm{DR}$ value indicates a smaller $r_1^{(m)}$ relative to $r_2^{(m)}$, suggesting higher correctness of the early prediction. Since $r_1^{(m)}$ and $r_2^{(m)}$ effectively reflect the correctness of early predictions, we focus on these two distances and ignore the distances between the sample and prototypes of other classes for efficiency.

\begin{algorithm}[!t]
	\caption{Inference Process of DE$^3$-BERT.}
    \label{algorithm:inference_process}
	\begin{algorithmic}
		\STATE \textbf{Input:} $M$-layer model, input sample $x$, and threshold $\tau$
		\STATE \textbf{Output:} Class probability distribution $p_x$
        \STATE $h_x=model.Embedding(x)$;
		\FOR{$i=1$ to $M$}
		    \STATE $h_x=model.EncoderBlocks[i](h_x)$;
            \STATE $p_x=model.Classifiers[i](h_x)$;
            \STATE \textcolor{green!50!black}{// Compute the entropy of $p_x$}
            \STATE $\mathrm{E_x}=Entropy(p_x)$;
            \STATE \textcolor{green!50!black}{// Map the hidden states to the metric space}
            \STATE $h'_x=model.PrototypicalNetworks[i](h_x)$;
            \STATE $r_{1}=CosineDistance(h'_x,{cp}_{k1})$;
            \STATE $r_{2}=CosineDistance(h'_x,{cp}_{k2})$;
            \STATE \textcolor{green!50!black}{// Compute the distance ratio between $r_1$ and $r_2$}
		    \STATE $\mathrm{DR_x}=DistanceRatio(r_1,r_2)$;
            \STATE \textcolor{green!50!black}{// Formulate the hybrid exiting indicator}
            \STATE $\mathrm{EDR_x}=HarmonicMean(\mathrm{E_x},\mathrm{DR_x})$;
		    \IF{$\mathrm{EDR_x} < \tau$}
		        \RETURN $p_x$;\hfill \textcolor{green!50!black}{// Exit at the $i$th layer}
		    \ENDIF
		\ENDFOR
		\RETURN $p_x$;
	\end{algorithmic}
\end{algorithm}

\subsubsection{Hybrid Exiting Strategy} 
In this paper, we propose a hybrid exiting strategy that integrates entropy and distance ratio.
Specifically, entropy and distance ratio can be regarded as two complementary measures of prediction correctness, derived from local and global information, respectively. 
Our objective is to enhance the estimation of prediction correctness by jointly considering both perspectives, thereby enabling more reliable and stable exiting decisions.
We argue that an early prediction can be deemed sufficiently correct only when the measures of prediction correctness from both local and global  perspectives are simultaneously high, i.e., when both entropy and distance ratio values are sufficiently low.
Notably, the harmonic mean has been widely adopted for integrating heterogeneous yet complementary measures. For instance, in the computation of the F1 score, precision and recall are fused via harmonic averaging to balance different dimensions and prevent the final decision from being excessively influenced by a single metric. Inspired by this principle, we devise a hybrid exiting indicator, $\rm EDR$, which computes the harmonic mean of entropy and distance ratio:
\begin{equation}\label{edr}
 {\rm EDR}=\frac{\lambda+1}{\frac{\lambda}{\rm{DR}}+\frac{1}{\rm Entropy}}
\end{equation}
where $\mathrm{DR}$ is the normalized distance ratio, and $\lambda$ is the fusion coefficient used to balance the two factors. The $\mathrm{EDR}$ value is always between 0 and 1. A lower $\mathrm{EDR}$ value indicates higher correctness of the early prediction from both global and local perspectives. Therefore, once the $\mathrm{EDR}$ value falls below the predefined threshold $\tau$, the inference process is terminated. The inference process of our DE$^3$-BERT is summarized in Algorithm~\ref{algorithm:inference_process}. For simplicity in expression, entropy and distance ratio mentioned in this paper refer to normalized values.

\section{Experiments}
\label{section:experiments}
In this section, we evaluate the DE$^3$-BERT framework's performance in accelerating inference on the GLUE benchmark. Additionally, we conduct ablation studies to demonstrate the effectiveness of each component within the framework.
We further validate the generality of our framework on different backbones.

\subsection{Tasks and Datasets}
We select six classification tasks from the GLUE benchmark~\cite{glue} for experiments, including SST-2, MRPC, QNLI, RTE, QQP, and MNLI. We exclude the STS-B task since it is a regression task. Besides, following the previous studies in~\cite{Berxit,globalpast,Deebert,zhu2021leebert}, we also exclude the WNLI and CoLA tasks. The dataset statistics for all tasks are listed in Table~\ref{tab:dataset_statistics}.

\begin{table}[!t]
\caption{Dataset statistics. NLI denotes the Natural Language Inference task, and QA denotes the Question Answering task.}% for the datasets used in our experiments
\label{tab:dataset_statistics}
\centering
\scalebox{0.9}{
  \begin{tabular}{lcccc}
    \toprule
    Dataset & Classes &  $|$Train$|$  & $|$Test$|$ &  Task  \\
    \midrule
    % CoLA & 2 &	 8.5k &	1k & Acceptability  \\
    SST-2 & 2 &  67k &  1.8k & Sentiment \\
    MRPC & 2 &  3.7k &  1.7k & Paraphrase \\
    QQP & 2 &  364k &  391k & Paraphrase \\
    MNLI & 3 &  393k &  20k & NLI \\
    QNLI & 2 &  105k &  5.4k & QA/NLI \\
    RTE & 2 &  2.5k &  3k & NLI \\
    \bottomrule
\end{tabular}}
\vspace{-3mm}
\end{table}

\subsection{Baselines}
We compare DE$^3$-BERT with four groups of baselines:

\begin{itemize}[leftmargin=*] 
\item{\it Backbone:}\,\, 
The original backbone serves as a reference for all inference acceleration methods, with a speed-up ratio of $1.0\times$.
We mainly choose the widely used BERT-base~\cite{bert} as the backbone model for convincing comparison. We also conduct experiments using RoBERTa-base~\cite{abs-1907-11692} on a representative subset of GLUE to evaluate the generality of our method across different backbones.

\item{\it Budget Exiting:}\,\, 
We directly train a BERT-base with $k$ layers and denote it as BERT-$k$L. We set $k$ to 6 and 4 to achieve $2.0\times$ and $3.0\times$ speed-up ratios, respectively. These two baselines force all samples to exit at a fixed layer, providing a lower bound for dynamic early exiting methods since no adaptive mechanisms are applied.

\item{\it Knowledge Distillation:}\,\, 
We select several classic knowledge distillation methods as references, including DistilBERT~\cite{distilbert}, PD-BERT~\cite{wellread}, BERT-PKD~\cite{PKD}, and BERT-of-Theseus~\cite{theseus}. These methods leverage different distillation strategies to compress a 12-layer BERT-base model into a 6-layer version, achieving a $2.0\times$ speed-up ratio.

\item{\it Early Exiting:}\,\, 
We primarily include two groups of early exiting methods for comparison. The first group encompasses all existing exiting strategies, which serves to validate the effectiveness of the proposed distance-enhanced hybrid exiting strategy.
It includes the confidence-based strategy DeeBERT~\cite{Deebert}, the patience-based strategy PABEE~\cite{Pabee}, the learnable strategy BERxiT~\cite{Berxit}, and the hybrid strategy PCEE-BERT~\cite{PCEE}.
The second group includes orthogonal works relevant to our study, i.e., GPFEE~\cite{globalpast}, LeeBERT~\cite{zhu2021leebert}, and COSEE~\cite{COSEE}, offering a benchmark for evaluating the performance of our DE$^3$-BERT within the broader early exiting community. Notably, different from our method which improves exiting strategies, these orthogonal works focus on improving the training schemes or architectures of internal classifiers, addressing early exiting issues from different perspectives. Hence, comparisons with these orthogonal works may not fully represent the effectiveness of the proposed distance-enhanced hybrid exiting strategy. 

For fair comparisons, HashEE~\cite{hashbased} is not included since it employs a more fine-grained token-level early exiting method, whereas our method focuses on sentence-level early exiting. CascadeBERT~\cite{li2021cascadebert} is also excluded from the comparison, as it implements early exiting within multiple cascaded complete models instead of a single model with multiple exits employed by our method. 
\end{itemize} 

More details about the above baselines can be found in Appendix A.

\subsection{Experimental Settings}
\label{section:experimental_settings}

\subsubsection{Speed Measurement} 
Since the actual runtime is unstable across different runs, following~\cite{PCEE,globalpast}, we manually adjust the threshold $\tau$ and measure the corresponding speed-up ratio based on the number of layers saved during inference:
\begin{equation}\label{speedup-ratio}
\text{Speed-up Ratio}=\frac{\sum_{m=1}^M M\times N^m}{\sum_{m=1}^M m\times N^m}
\end{equation}
where $M$ is the total number of layers and $N^m$ is the number of samples exiting from the $m$th layer.
We confirm the rationale for this speed measurement in Appendix C-B.

\subsubsection{Training} 
Our implementation is based on Hugging Face's Transformers~\cite{huggingface}. Both the internal classifiers and the prototypical networks are composed of a single linear layer. We choose Center Loss as DAR if not specified and cosine distance as the distance metric. Following the previous work~\cite{Pabee,PCEE,globalpast}, we perform a grid search over learning rates of \{1e-5, 2e-5, 3e-5, 5e-5\}, batch sizes of $\{16, 32, 128\}$, and $\alpha$ values (i.e., the regularization coefficient defined in Equation~(\ref{loss-m-1})) of $\{0.0001, 0.001, 0.01, 0.1\}$. We set the $\gamma$ value in Equation~(\ref{cp-update}) as $0.5$. The maximum sequence length is fixed at $128$. We employ a linear decay learning rate scheduler and the AdamW~\cite{LoshchilovH19} optimizer. We conduct experiments on one single RTX3090 GPU with 24GB.

\subsubsection{Inference}
Following the previous work~\cite{PCEE,globalpast}, we use a batch size of $1$ during inference, which simulates a common industry scenario where requests from different users arrive one by one. We select the fusion coefficient $\lambda$ (in Equation~(\ref{edr})) from \{0.667, 1.0, 1.5, 2.0, 3.0\} for each task.
For fair comparisons, we manually adjust the threshold $\tau$ for each task to achieve similar speed-up ratios as the baseline methods, i.e., $2.0\times$ and $3.0\times$, and then compare their task performance under these two speed-up ratios.

\begin{table*}[!t]
% \vspace{-2mm}
\caption{Comparison with dynamic early exiting methods on the GLUE test sets at speed-up ratios of 2.0$\times$ and 3.0$\times$. The $\ddag$ denotes the results from our implementation, while the $\dag$ denotes the results from the original paper. Other baseline results are taken from GPFEE~\cite{globalpast}. DE$^3$-BERT uses the Center Loss as DAR. We report the performance score for each task, with the corresponding speed-up ratios shown in parentheses. The - denotes unavailable results. Best performance values are marked in bold. The * denotes a statistically significant performance improvement over the competitive baselines PCEE-BERT and GPFEE, while the $\circ$ denotes a statistically significant improvement over the state-of-the-art baseline COSEE, both at a significance level of 0.05.}
\label{tab:main_result}
\centering
\scalebox{0.9}{
\begin{tabular}{clccccccl}
\toprule
&\multirow{2}{0.1cm}{Method} &   MNLI  & MRPC  &  QNLI  &  QQP  &  RTE  &  SST-2 & \multirow{2}{1.0cm}{AVG}\\
% &  (393K)& (3.7K)&  (105K)& (364K)& (2.5K)& (67K)  & \\
&& Acc & F1/Acc  & Acc  & F1/Acc   & Acc   & Acc  & \\
\midrule
&BERT-base$^\dag$ &84.6 (1.00$\times$) & 88.9/- (1.00$\times$)& 90.5 (1.00$\times$) & 71.2/- (1.00$\times$)& 66.4 (1.00$\times$) & 93.5 (1.00$\times$) &-\\
 \midrule
\multirow{10}{0.1cm}{\rotatebox{90}{$\sim$2$\times$}}&
BERT-6L & 
     80.8 (2.00$\times$) & 85.1/78.6 (2.00$\times$)& 
     86.7 (2.00$\times$) & 69.7/88.3 (2.00$\times$) &
    63.9 (2.00$\times$) & 
    91.0 (2.00$\times$) &
    80.5\\
&DeeBERT & 
    74.4 (1.87$\times$) & 
    84.4/77.4 (2.07$\times$) & 
    85.6 (2.09$\times$) & 
    70.4/88.8 (2.13$\times$) &
    64.3 (1.95$\times$) & 
    90.2 (2.00$\times$) &
    79.2\\
&PABEE & 
    79.8 (2.07$\times$) & 
    84.4/77.4 (2.01$\times$) & 
    88.0 (1.87$\times$) & 
    70.4/88.6 (2.09$\times$) &
    64.0 (1.81$\times$) & 
    89.3 (1.95$\times$) &
    80.3\\
&{BERxiT}$^\ddag$  & 	
    74.5 (2.06$\times$) & 
    84.3/77.2 (1.99$\times$)& 
    85.5 (1.98$\times$)&	
    70.2/88.5 (2.01$\times$)&	
    64.2 (1.97$\times$)& 	
    90.2 (1.98$\times$) &
    79.1\\
&PCEE-BERT$^\ddag$ &	
    81.0 (1.98$\times$)& 	  86.1/80.2 (2.02$\times$)&
    87.9 (2.04$\times$)& 	
    70.8/89.1 (2.02$\times$)&	
    {\bf 65.7} (1.99$\times$)&	
    91.0 (2.04$\times$)&
    81.5 \\
&GPFEE$^\dag$ & 
    {\bf 83.3} (1.96$\times$) & 
    87.0/81.8 (1.98$\times$) & 
    89.8 (1.97$\times$) & 
    {\bf 71.2/89.4} (2.18$\times$) &
    64.5 (2.04$\times$) & 
    {\bf 92.8} (2.02$\times$)&
    82.5 \\
&LeeBERT$^\dag$ &	
    83.1 (1.97$\times$)& 	
    {\bf 87.1/-} (1.97$\times$)&	
    - & 	
    - &	
    - &	
    92.6 (1.97$\times$) &-\\
&{\bf DE$^3$-BERT (ours)} &	
    83.2 (2.04$\times$)& 	
    86.6/81.5 (1.98$\times$)&	
    {\bf 90.0}$^*$ (2.07$\times$)& 	
    {\bf 71.2/89.4} (2.16$\times$)&	
    {\bf 65.7} (1.99$\times$)&	
    92.5 (2.02$\times$)&
    {\bf 82.6}$^*$ \\
\\[-0.25cm]
\cline{2-9}
\\[-0.25cm]
&COSEE$^\dag$ &	
    83.4 (1.92$\times$)& 	
    88.0/82.0 (2.70$\times$)&	
    90.2 (2.56$\times$)& 	
    {\bf 71.4/89.4} (2.01$\times$)&	
    68.7 (1.96$\times$)&	
    {\bf 93.0} (2.14$\times$)&
    83.5 \\
&{\bf DE-COSEE (ours)} &	
    {\bf 83.5}$^\circ$ (1.94$\times$)& 	
    {\bf 88.6/82.2}$^\circ$ (2.68$\times$)&	
    {\bf 90.3}$^\circ$ (2.51$\times$)& 	
    {\bf 71.4/89.4} (2.05$\times$)&	
    {\bf 68.9}$^\circ$ (1.98$\times$)&	
    {\bf 93.0} (2.14$\times$)&
    \bf{83.6}$^\circ$ \\
 \midrule
\multirow{9}{0.1cm}{\rotatebox{90}{$\sim$3$\times$}}&BERT-4L & 
     77.6 (3.00$\times$) & 82.9/74.9 (3.00$\times$)& 
     85.4 (3.00$\times$) & 67.7/87.5 (3.00$\times$) &
    63.0 (3.00$\times$) & 
    88.7 (3.00$\times$) & 78.5\\
&DeeBERT & 
    61.0 (2.80$\times$) & 
    83.5/75.5 (2.61$\times$) & 
    80.8 (2.88$\times$) & 
    66.1/86.9 (3.19$\times$) &
    60.5 (2.90$\times$) & 
    84.7 (2.71$\times$) &73.8\\
&PABEE & 
    75.9 (2.70$\times$) & 
    82.6/73.1 (2.72$\times$) & 
    82.6 (3.04$\times$) & 
    69.5/88.2 (2.57$\times$) &
    60.5 (2.38$\times$) & 
    85.2 (3.15$\times$) &76.8\\
&BERxiT$^\ddag$ & 	
    60.9 (3.01$\times$) & 
    83.4/75.4 (3.06$\times$) & 
    80.3 (2.89$\times$)&	
    64.3/86.2 (3.14$\times$)&	
    60.7 (2.87$\times$)& 	
    84.5 (2.86$\times$)  &73.5\\
&PCEE-BERT$^\ddag$ &	
    77.8 (3.08$\times$)& 	
    82.8/74.3 (2.98$\times$)&	
    85.6 (3.06$\times$)& 	
    69.6/88.3 (2.96$\times$)&	
    63.1 (2.89$\times$)&	
    88.9 (2.92$\times$)&
    78.8 \\
&GPFEE$^\dag$ & 
    78.4 (2.99$\times$) & 
    {\bf 84.5/77.7} (2.87$\times$) & 
    {\bf 87.3} (2.78$\times$) & 
    70.4/89.2 (3.16$\times$) &
    63.0 (2.88$\times$) & 
    91.1 (2.97$\times$)& 80.1\\
&{\bf DE$^3$-BERT (ours)} &	
    {\bf 79.9}$^*$ (2.99$\times$)& 	
    83.8/76.5 (3.01$\times$)&	
    87.0 (3.02$\times$)& 	
    {\bf 70.6/89.3}$^*$ (3.18$\times$)&	
    {\bf 63.6}$^*$ (2.97$\times$)&	
    {\bf 91.4}$^*$ (2.98$\times$)&  {\bf 80.3}$^*$\\
\\[-0.25cm]
\cline{2-9}
\\[-0.25cm]
&COSEE$^\ddag$ &	
    80.0 (2.98$\times$) & 
    84.2/76.9 (3.05$\times$) & 
    87.2 (3.03$\times$) & 
    70.8/89.4 (3.18$\times$) &
    64.1 (2.98$\times$) & 
    91.7 (2.97$\times$)& 80.6\\
&{\bf DE-COSEE (ours)} &	
    {\bf 80.3}$^\circ$ (3.02$\times$) & 
    {\bf 84.9/77.9}$^\circ$ (3.03$\times$) & 
    {\bf 87.5}$^\circ$ (3.01$\times$) & 
    {\bf 71.2/89.4}$^\circ$ (3.17$\times$) &
    {\bf 64.9}$^\circ$ (2.99$\times$) & 
    {\bf 92.0}$^\circ$ (3.02$\times$)& \bf{81.1}$^\circ$\\
\bottomrule
\end{tabular}}
\vspace{-1mm}
\end{table*}

\begin{table}[!t]
\caption{Comparison with knowledge distillation methods on the test sets of four GLUE tasks. All distilled models listed below have six layers and the speed-up ratio is approximately 2.0$\times$ (±8$\%$). We report accuracy for all tasks. $\ddag$ denotes the results taken from MobileBERT~\cite{sun2020mobilebert}. Other baseline results are taken from BERT-of-Theseus~\cite{theseus}. Best performance values are marked in bold. The * denotes a statistically significant performance improvement over the baselines PD-BERT and BERT-of-Theseus at a significance level of 0.05.}
\label{tab:compare_static_methods}
\centering
\scalebox{0.9}{
\begin{tabular}{cllllll}
\toprule
    & Method & MNLI &   QNLI &  QQP & SST-2     &  AVG \\
    \midrule
   \multirow{5}{0.05cm}{\rotatebox{90}{$\sim$2$\times$}}& DistilBERT$^\ddag$   &  
   {81.9} &	88.2 &	
   88.4 &	
   92.1 & 87.7 \\
    & PD-BERT        &  
    {82.8} &	88.9 &	88.9 &	91.8 & 88.1  \\
    & BERT-PKD        &  81.5 &	89.0 &	88.9 &	92.0 & 87.9  \\
    & BERT-of-Theseus &  82.4 	& 89.6 	& 89.3 	& {92.2}  & 88.4   \\
     & \bf{DE$^3$-BERT} (ours)         &  {\bf{83.2}}$^*$  &   {\bf{90.0}}$^*$  &  {\bf{89.4}}$^*$  &   {\bf 92.5}$^*$   &  \bf{88.8}$^*$   \\
    \bottomrule
\end{tabular}}
\end{table}

\subsection{Overall Performance Comparison}
In Table~\ref{tab:main_result}, we report the classification performance of each early exiting method on the GLUE test sets at speed-up ratios of 2.0$\times$ and 3.0$\times$, respectively.
Regarding similar research that improves exiting strategies during inference, namely BERT-$k$L, DeeBERT~\cite{Deebert}, PABEE~\cite{Pabee}, BERxiT~\cite{Berxit}, and PCEE-BERT~\cite{PCEE}, our DE$^3$-BERT consistently outperforms these baseline methods across all speed-up ratios by a clear margin, validating the effectiveness of the proposed distance-enhanced hybrid exiting strategy. 
Specifically, as the speed-up ratio increases, the performance of these baselines drops dramatically, while the superiority of our method becomes more significant. For instance, at speed-up ratios of 2.0$\times$ and 3.0$\times$, our DE$^3$-BERT outperforms the competitive baseline PCEE-BERT~\cite{PCEE} by an average of 1.1 and 1.5 points across all tasks, respectively. This indicates that by incorporating distance-based global information, our framework effectively improves the reliability of exiting decisions, yielding a better performance-efficiency trade-off for PLMs.
Additionally, regarding the remaining orthogonal works that enhance the training schemes or architectures of internal classifiers, our DE$^3$-BERT yields comparable results to GPFEE~\cite{globalpast} and LeeBERT~\cite{zhu2021leebert}. Since the state-of-the-art baseline COSEE~\cite{COSEE} utilizes a calibrated sample weighting mechanism in formulating its training objective, while our DE$^3$-BERT does not employ this technique, a direct comparison would be unfair. To address this issue, we develop a variant DE-COSEE (Distance-Enhanced COSEE), which combines COSEE’s training method with our proposed distance-enhanced hybrid exiting strategy to ensure a fair comparison with the baseline COSEE. The experimental results show that DE-COSEE consistently surpasses the original COSEE in nearly all cases, confirming the effectiveness of our proposed distance-enhanced hybrid exiting strategy for early exiting networks trained with different methods.
Furthermore, the one-tailed t-tests verify the statistical significance of our method's improvements over the top-3 baselines PCEE-BERT~\cite{PCEE}, GPFEE~\cite{globalpast}, and COSEE~\cite{COSEE} in most cases, providing strong evidence for the superiority of our method in accelerating the inference of PLMs.
Notably, our method introduces minimal additional computational and storage costs (see Appendix~C-G).
Besides, we further explore the impact of hyperparameters and different DAR forms (see Appendices~C–E and~C–J), and investigate the generality of our method on out-of-distribution (OOD) data and Chinese datasets (see Appendices~C–C and~C–D).

We also compare DE$^3$-BERT with several classic knowledge distillation methods in Table~\ref{tab:compare_static_methods}. The experimental results consistently demonstrate the superiority of our method, and the one-tailed t-tests further confirm the statistical significance of our method's improvements over the competitive baselines PD-BERT~\cite{wellread} and BERT-of-Theseus~\cite{theseus} across all tasks.
We attribute this to the static inference process of knowledge distillation methods, where samples of varying complexities pass through the entire model indiscriminately. Instead, our method establishes a reliable dynamic inference process based on sample complexity by incorporating class prototypes. This effectively avoids redundant computations and further enhances the performance-efficiency trade-off of PLMs.

\subsection{Ablation Studies}

\subsubsection{Effect of Distance Ratio} 
\begin{figure}[!t]
\centering
\vspace{-7mm}
\subfloat[QQP]{\includegraphics[width=0.46\linewidth]{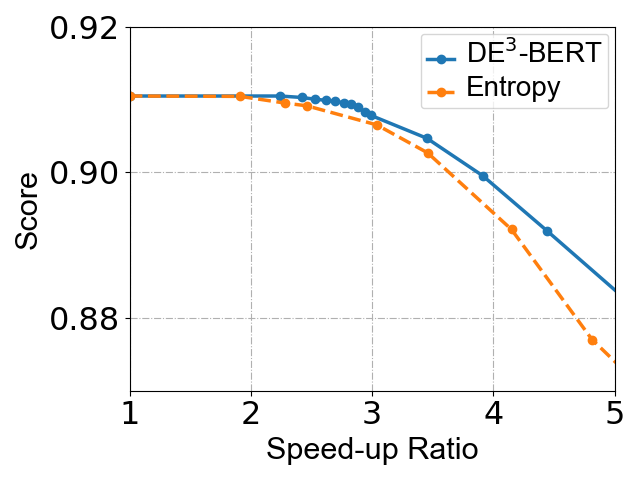}%
\label{fig:ablation_dr_qqp}}
\hfil
\subfloat[SST-2]{\includegraphics[width=0.46\linewidth]{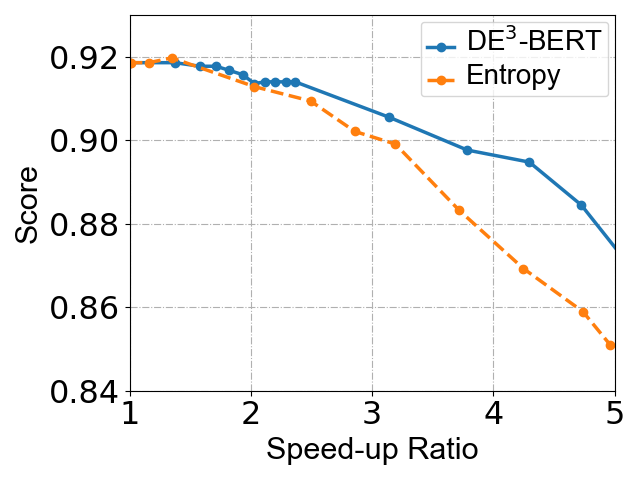}%
\label{fig:ablation_dr_sst2}}
\hfil
\vspace{-3mm}
\subfloat[MNLI]{\includegraphics[width=0.46\linewidth]{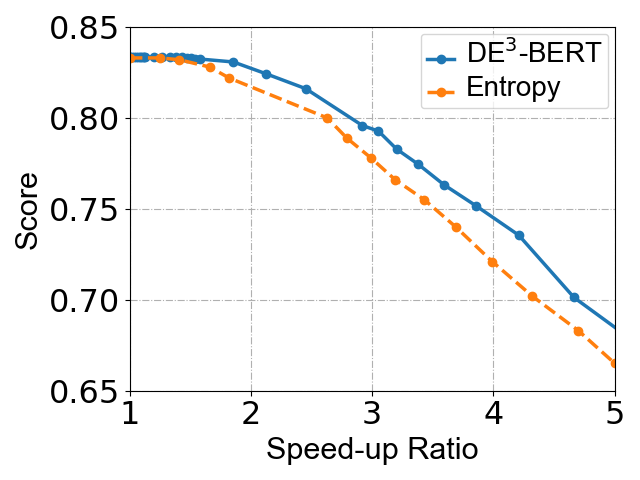}%
\label{fig:ablation_dr_mnli}}
\hfil
\subfloat[QNLI]{\includegraphics[width=0.46\linewidth]{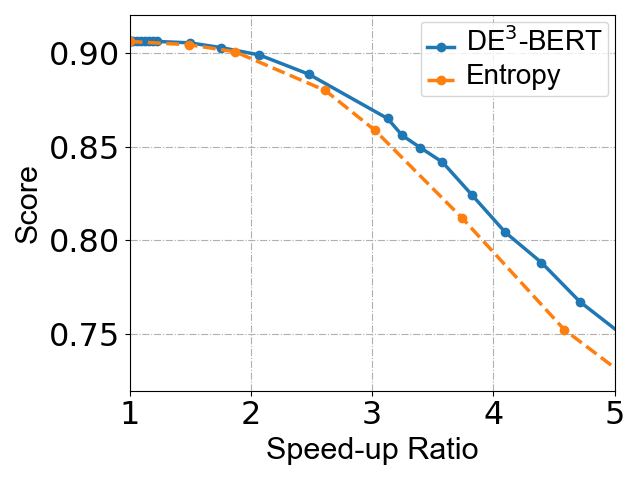}%
\label{fig:ablation_dr_qnli}}
\caption{Performance-efficiency trade-off curves using different exiting strategies on the development sets of four GLUE tasks. DE$^3$-BERT is our hybrid exiting strategy. Entropy denotes the classic entropy-based exiting strategy. Each point on the curve corresponds to a selected threshold. Its horizontal and vertical coordinates represent the corresponding speed-up ratio and task performance, respectively. Models are trained as illustrated in Section \ref{section:training_scheme}.}%(a) QQP. (b) SST-2. (c) MNLI. (d) QNLI.
\label{fig:ablation_dr}
\vspace{-1mm}
\end{figure}

As mentioned in Section \ref{section:overview}, our DE$^3$-BERT leverages the global information (distance ratio) to enhance the entropy-based early exiting. To further investigate the effect of distance ratio in exit decision-making, we conduct ablation studies on a representative subset of GLUE to compare the entropy-based exiting strategy with our hybrid exiting strategy. Note that these exiting strategies are applied to the same model trained as illustrated in Section \ref{section:training_scheme} for each task. Consequently, the experimental results encompass the performance improvements brought by DAR, and our focus is solely on the effect of different exiting strategies employed during inference. Fig.~\ref{fig:ablation_dr} shows the performance-efficiency trade-off curves of different exiting strategies. As we can see, our hybrid exiting strategy consistently outperforms the entropy-based strategy by a clear margin under various speed-up ratios across all tasks, confirming the effectiveness of incorporating the distance ratio into the exit decision-making process. This suggests that the distance-based global information effectively complements the entropy-based local information, thus enhancing the reliability of exiting decisions for better performance-efficiency trade-offs. In addition, it is noticeable that the performance improvements brought by the distance ratio appear to be more significant under high speed-up ratios, demonstrating the advantage of DE$^3$-BERT in high acceleration scenarios. We subsequently analyze the reasons for this observation through experiments (see Appendix~C-F).

\subsubsection{Effect of DAR and Prototypical Networks}  
\begin{figure}[!t]
\centering
\vspace{-3mm}
\subfloat[QQP]{\includegraphics[width=0.46\linewidth]{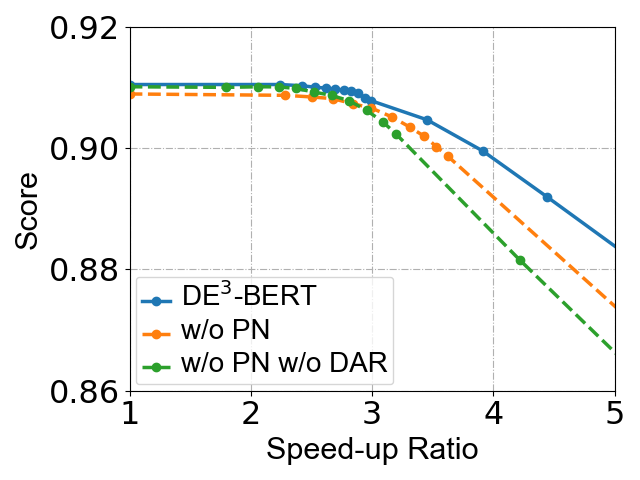}%
\label{fig:ablation_dar_pn_qqp}}
\hfil
\subfloat[SST-2]{\includegraphics[width=0.46\linewidth]{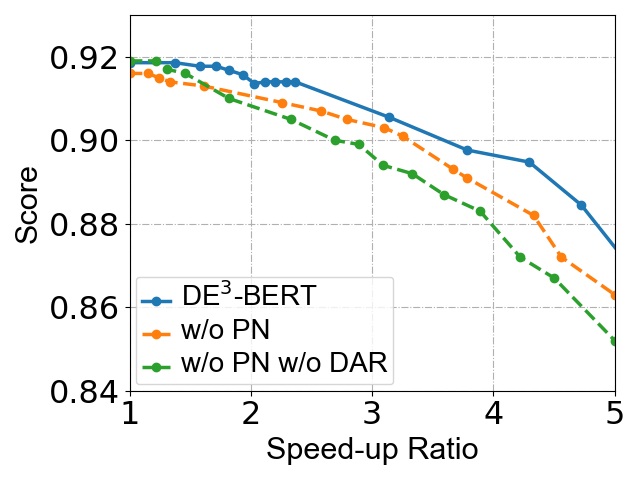}%
\label{fig:ablation_dar_pn_sst2}}
\hfil
\vspace{-3mm}
\subfloat[MNLI]{\includegraphics[width=0.46\linewidth]{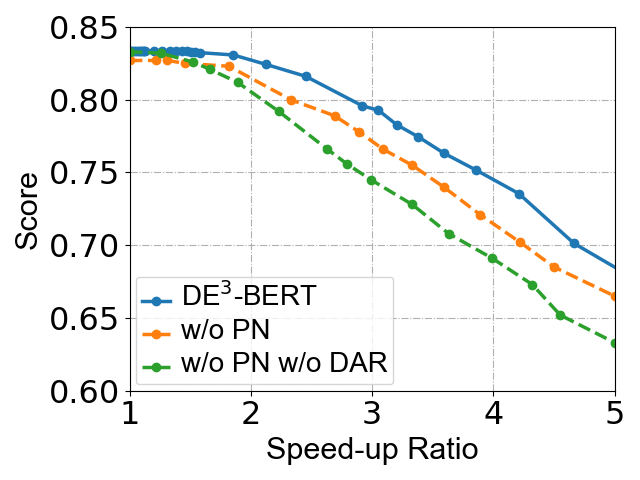}%
\label{fig:ablation_dar_pn_mnli}}
\hfil
\subfloat[QNLI]{\includegraphics[width=0.46\linewidth]{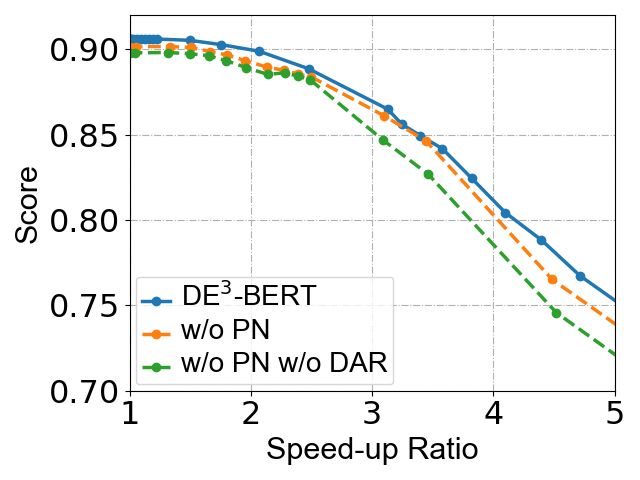}%
\label{fig:ablation_dar_pn_qnli}}
\caption{Impact of prototypical networks and DAR on the trade-off between model performance and efficiency on the development sets of four GLUE tasks. We employ the proposed hybrid exiting strategy for all models during inference. PN denotes the prototypical networks.}% (a) QQP. (b) SST-2. (c) MNLI. (d) QNLI.
\label{fig:ablation_dar_pn}
\vspace{-3mm}
\end{figure}

To investigate the effect of DAR and the prototypical networks, we plot the performance-efficiency trade-off curves of different models on a representative subset of GLUE, as shown in Fig.~\ref{fig:ablation_dar_pn}. All models adopt the proposed hybrid exiting strategy during inference. Implementation details can be found in Appendix B.
We observe that removing either DAR or the prototypical networks causes significant performance degradation, demonstrating their effectiveness in learning high-quality class prototypes and metric spaces. This is crucial for accurately estimating the prediction correctness and making reliable exiting decisions.
We further visualize the effect of DAR in learning high-quality metric space and class prototypes (see Appendix~C-H). Additionally, we reveal that the prototypical networks can prevent the homogenization of entropy and distance ratios, facilitating their mutual correction (see Appendix~C-I).

\subsection{Generality on Different PLMs}
\label{section:generality}
To investigate the generality of our method on different PLMs, we conduct experiments on a representative subset of GLUE with RoBERTa~\cite{abs-1907-11692} as the backbone model, as shown in Table~\ref{tab:generalization}. 
RoBERTa is an optimized variant of BERT that utilizes more training data, larger batch sizes, and better training techniques.
The experimental results indicate that our method generally outperforms all early exiting baselines, and the one-tailed t-tests further verify the statistical significance of our method's improvements over the competitive baseline GPFEE~\cite{globalpast} in most tasks. These findings demonstrate the generality of our framework across different PLMs.

\begin{table}[!t]
\vspace{-3mm}
\caption{Comparison with dynamic early exiting methods on the test sets of four GLUE tasks with RoBERTa as the backbone. The speed-up ratio is approximately 3.0$\times$ (±11$\%$). We report F1-score for QQP and accuracy for other tasks. $\dag$ denotes the results taken from the original paper. Other baseline results are taken from CascadeBERT~\cite{li2021cascadebert}. The - denotes unavailable results of PABEE. Best performance values are marked in bold. The * denotes a statistically significant performance improvement over the competitive baseline GPFEE at a significance level of 0.05.}
\label{tab:generalization}
\centering
\scalebox{0.9}{
\begin{tabular}{llllll}
\toprule
Method & MNLI &   QNLI &  QQP & SST-2     &  AVG \\
    \midrule
RoBERTa-base & 
     87.0 & 
     92.4 & 
     71.8 & 
     94.3 & 
     86.4 \\
     \midrule
RoBERTa-4L & 80.3 &	86.2 & 69.8	 &	90.8 & 81.8        \\
DeeBERT         &  53.9 &	77.2 &	67.6 &	88.6 & 71.8 \\
PABEE        &  74.0 &	- &	- &	87.5 & - \\
GPFEE$^\dag$ &  81.4	& {89.2}	& 
    {\bf 71.9}	& 
    {93.5} & 84.0   \\
\bf{DE$^3$-RoBERTa (ours)}           &  
     {\bf 83.1}$^*$ &  
     {\bf 89.3}$^*$ &  
     71.6 &   
     {\bf 93.7}$^*$  &  \bf{84.4}$^*$   \\
    \bottomrule
\end{tabular}}
\vspace{-3mm}
\end{table}

\section{Conclusion}
\label{section:conclusion}
In this article, we point out that the performance limitation of existing early exiting methods primarily lies in their excessive focus on local information from individual test samples while neglecting the global information offered by the sample population, which compromises the estimation of prediction correctness, leading to erroneous exiting decisions. To remedy this, we propose the DE$^3$-BERT framework, which leverages prototypical networks to provide distance-based global information, improving the estimation of prediction correctness for reliable early exiting. Our framework is both intuitive and interpretable. 
Extensive experiments on the GLUE benchmark demonstrate the superiority of our framework under various speed-up ratios with minimal additional storage and computational costs. Further analysis validates the effectiveness and interpretability of each component in our framework, as well as its robustness to hyperparameter selection and its strong generality across out-of-distribution (OOD) data, different languages, and various backbone models.
We also provide a detailed discussion of the experimental results, limitations, scalability, and real-world applications and implications of our framework (see Appendix~E).

\bibliography{reference}
\bibliographystyle{IEEEtran}

% \clearpage
\appendices

\section{Baseline Competitors}
\label{section:baselines}
In this section, we introduce the comparative baselines in our experiments in detail.

\subsubsection{Backbone}
BERT-base~\cite{bert} consists of 12 encoders with a single-layer classifier attached to the last encoder. With 110M parameters, a hidden size of 768, and 12 attention heads per layer, it is first pre-trained on the Wiki corpus and then fine-tuned on specific tasks.

\subsubsection{Budget Exiting}
BERT-$k$L preserves the first $k$ encoders of the BERT-base with a single-layer classifier attached on top. It is initialized with pre-trained weights from the first $k$ layers of the BERT-base and then fine-tuned on specific tasks. BERT-4L and BERT-6L essentially employ static early exiting strategies, where all samples exit early at a fixed layer. Hence, these models serve as a lower bound for comparison with dynamic early exiting methods.

\subsubsection{Knowledge Distillation}
Knowledge distillation methods compress models by transferring knowledge from a larger teacher model to a more compact student model. DistilBERT~\cite{distilbert} is distilled during the pre-training phase. PD-BERT~\cite{wellread} is first pre-trained on a corpus and then distilled during the fine-tuning stage. BERT-PKD~\cite{PKD} is distilled using both the final predictions and intermediate-layer outputs provided by the teacher model during the fine-tuning stage. BERT-of-Theseus~\cite{theseus} is distilled using a progressive module replacement strategy during the fine-tuning stage. All of the compressed models mentioned above consist of 6 encoders, resulting in a speed-up ratio of 2.0$\times$.

\subsubsection{Early Exiting} 
Regarding the inference stage, DeeBERT~\cite{Deebert}, GPFEE~\cite{globalpast}, and COSEE~\cite{COSEE} employ confidence-based exiting strategies, which use the entropy or energy value of early predictions to reflect the confidence level of internal classifiers. The inference process is terminated once the entropy or energy value falls below a predefined threshold. PABEE~\cite{Pabee} and LeeBERT~\cite{zhu2021leebert} both employ a patience-based exiting strategy, which utilizes cross-layer consistency as the exiting indicator. The exiting criterion is met when there are enough (i.e., meeting the patience threshold) consecutive internal classifiers that agree with each other. BERxiT~\cite{Berxit} incorporates a learning-to-exit network to score the correctness of early predictions. Early exit is triggered once the correctness score surpasses the threshold.
PCEE-BERT~\cite{PCEE} employs a hybrid exiting strategy that combines confidence-based and patience-based strategies. It uses entropy as the confidence measure and terminates the inference if there are enough (i.e., meeting the patience threshold) consecutive internal classifiers that are confident in their predictions.

Regarding the architecture of internal classifiers, GPFEE~\cite{globalpast} enhances the internal classifiers by integrating both past and future states as its inputs to facilitate high-quality early predictions, while the remaining models utilize a single-layer fully-connected network as the internal classifier.

Regarding the training phase, DeeBERT~\cite{Deebert}, PABEE~\cite{Pabee}, and PCEE-BERT~\cite{PCEE} are trained by minimizing the cross-entropy losses for all classifiers. For the remaining models, despite the cross-entropy loss for internal classifiers, additional training objectives are needed. BERxiT~\cite{Berxit} introduces mean squared error (MSE) for the learning-to-exit network to encourage high-quality scoring results for prediction correctness. GPFEE~\cite{globalpast} leverages the imitation loss for generating approximations of future states based on all available past states. LeeBERT~\cite{zhu2021leebert} leverages a cross-layer distillation objective with learned weights to encourage mutual learning among internal classifiers.  
COSEE~\cite{COSEE} introduces the online signal calibration loss to produce highly discriminative exiting indicators, facilitating the reliability of exiting decisions.

The above information is summarized in Table~\ref{tab:early_exit_baselines}.

\begin{table*}[ht]
\caption{Details of dynamic early exiting baselines. $\tau$ denotes the predefined threshold. IC denotes the internal classifier. FC denotes the fully-connected network. $\mathscr{L}_{\text{CE}}$ denotes the cross-entropy loss. $\mathscr{L}_{\text{COS}}$ denotes the imitation loss. $\mathscr{L}_{\text{KD}}$ denotes the distillation loss. $\mathscr{L}_{\text{OSC}}$ denotes the online signal calibration loss.}
\label{tab:early_exit_baselines}
\centering
  \begin{tabular}{lcccc}
    \toprule
    Method & Exiting Strategy &  IC Architecture & Exiting Criterion  & Training Objective  \\
    \midrule
    DeeBERT & Confidence-based &	Single-layer FC  &	${\rm Entropy}<\tau$ & $\mathscr{L}_{\text{CE}}$  \\
    PABEE & Patience-based &  Single-layer FC &  ${\rm PatienceCounter} \geq\tau$ & $\mathscr{L}_{\text{CE}}$ \\
    BERxiT & Learnable &  Single-layer FC &  ${\rm CorrectnessScore}>\tau$ & $\mathscr{L}_{\text{CE}}$, MSE \\
    % HashEE & Hash-based &  - &  ${\rm LayerCounter} \geq\text{Assigned Layer}$ & $\mathscr{L}_{\text{CE}}$ \\   
    PCEE-BERT & Hybrid & Single-layer FC &  ${\rm PatienceCounter}\geq \tau$ & $\mathscr{L}_{\text{CE}}$ \\
    GPFEE & Confidence-based &  Enhanced &  ${\rm Entropy}<\tau$ & $\mathscr{L}_{\text{CE}}$, $\mathscr{L}_{\text{COS}}$ \\
    % CascadeBERT & Confidence-based &  Enhanced &  ${\rm LabelScore}<\tau$ & $\mathscr{L}_{\text{CE}}$, Regularization \\
    LeeBERT & Patience-based &  Single-layer FC &  ${\rm PatienceCounter} \geq\tau$ & $\mathscr{L}_{\text{CE}}$, $\mathscr{L}_{\text{KD}}$ \\
    COSEE & Confidence-based &  Single-layer FC &  ${\rm Energy}<\tau$ & $\mathscr{L}_{\text{CE}}$, $\mathscr{L}_{\text{OSC}}$ \\
    \bottomrule
\end{tabular}
\end{table*}

\section{Implementation Details for Ablation Studies}
\label{section:implementation_details}
To verify the effectiveness of the prototypical networks (PNs) and DAR, we train two variants of our DE$^3$-BERT model and employ the proposed hybrid exiting strategy for these models as well as the full DE$^3$-BERT model during inference. We then compare the performance-efficiency trade-offs of different models.

\subsubsection{DE$^3$-BERT w/o PN}
To demonstrate the effectiveness of the prototypical networks, we design and train a variant of the DE$^3$-BERT model (denoted by DE$^3$-BERT w/o PN) as illustrated in Section III-B in the main paper. During inference, we use the proposed hybrid exiting strategy. Note that the computation of DAR terms and the updating of class prototype representations during training, as well as the calculation of distance ratios during inference, are all based on the sample representations obtained by pooling the hidden states of encoder layers (i.e., inputs to internal classifiers).

\subsubsection{DE$^3$-BERT w/o PN w/o DAR}
To further demonstrate the effectiveness of DAR, we also train another variant of the DE$^3$-BERT model (denoted by DE$^3$-BERT w/o PN w/o DAR) only using the cross-entropy loss as the training objective. To ensure the feasibility of the proposed hybrid exiting strategy during inference, the class prototype representations are still updated during training. Note that the updating of class prototype representations during training, as well as the calculation of distance ratios during inference, are all based on the sample representations obtained by pooling the hidden states of encoder layers (i.e., inputs to internal classifiers).

\section{Additional Experimental Results}
\subsection{Distribution of Exiting Layers}
\label{section:sample_distribution}
\begin{figure}[t]
\vspace{-5mm}
\centering
\subfloat[$\tau$=0.14 (0.906,1.34$\times$)]{\includegraphics[width=0.45\linewidth]{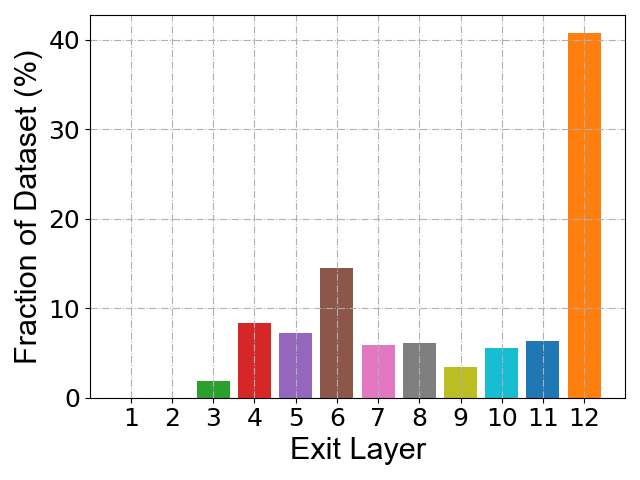}%
\label{fig:sample_distribution_1}}
\hfil
\subfloat[$\tau$=0.40 (0.901,2.07$\times$)]{\includegraphics[width=0.45\linewidth]{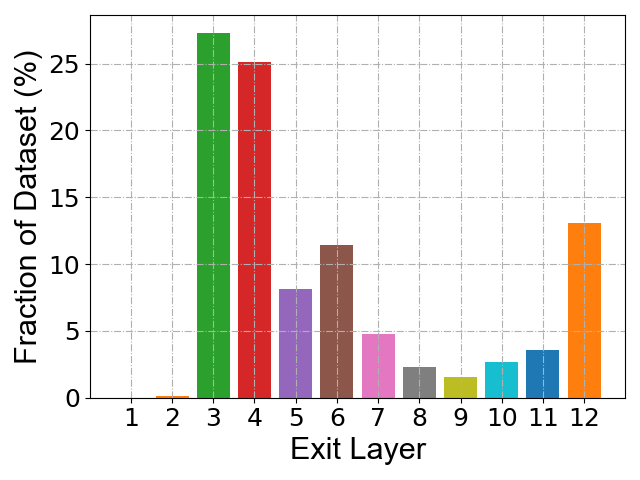}%
\label{fig:sample_distribution_2}}
\hfil
\subfloat[$\tau$=0.65 (0.811,4.04$\times$)]{\includegraphics[width=0.45\linewidth]{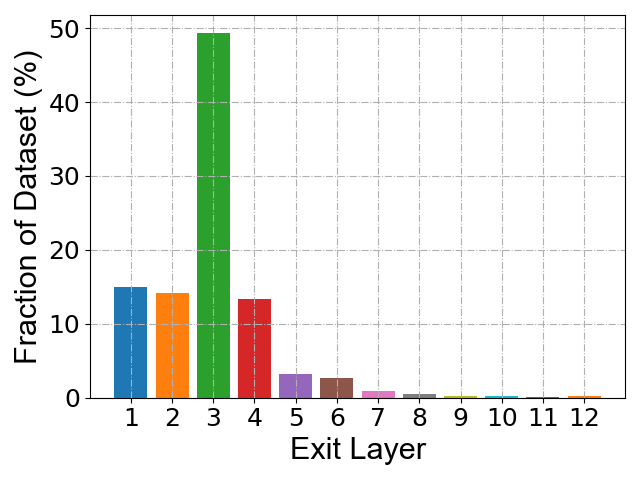}%
\label{fig:sample_distribution_3}}
\caption{Distribution of exiting layers on the QNLI development set with threshold settings of 0.14, 0.40, and 0.65. $\tau$ denotes the threshold, with corresponding task performance and speed-up ratios provided in parentheses.}
\label{fig:sample_distribution}
\end{figure}

To intuitively analyze the distribution of the sample's exiting layers for the DE$^3$-BERT framework, we calculate the proportion of samples exiting at different layers across various thresholds on the QNLI task, as shown in Fig.~\ref{fig:sample_distribution}. It can be observed that as the threshold increases, the samples tend to exit from shallow layers, resulting in a higher speed-up ratio, as well as a certain degree of performance degradation. This is consistent with our intuitive understanding.

\subsection{An Analysis of Different Speed Measurements}
\label{section:different_speed_measurements}
As mentioned in Section IV-C in the main paper, since the actual inference time is unstable across different runs, we calculate the speed-up ratio based on the number of executed layers during forward propagation as illustrated in Equation~(13) in the main paper. To validate the rationality of this speed measurement, we manually adjust the threshold during inference and collect the corresponding three types of speed measurements, including the total number of executed layers, the actual inference time, and the computational costs (FLOPs), on the SST-2 development set, as shown in Table~\ref{tab:different_speed_measurement}. We can observe that the total number of executed layers is approximately proportional to the model's computational costs, with a high Pearson correlation coefficient of 0.99 between them. This is because the encoder block constitutes the majority of the model's computational complexity, as depicted in Table~\ref{tab:flops_analysis}.
Additionally, the number of executed layers is highly positively correlated with the actual inference time, with a Pearson coefficient of 0.96. Therefore, the above analysis validates the rationality of the speed measurement based on the number of executed layers.

\begin{table}[t]
\caption{Speed measurements through the number of executed layers, inference time, and FLOPs on the SST-2 development set across various thresholds.}
\label{tab:different_speed_measurement}
\centering
  \begin{tabular}{cccc}
    \toprule
Threshold & Executed Layers & Inference Time (s) & FLOPs ($\times10^{12}$) \\
    \midrule
    0.000 & 10464 &	 17.07 &	19.0  \\
    0.001 & 10159 &  16.90 &  18.4 \\
    0.007 & 7118 &  15.24 &  12.9  \\
    0.010 & 6155 &  13.66 &  11.1  \\
    0.070 & 4491 &  12.64 &  8.1 \\
    0.140 & 3751 &  12.19 &  6.8 \\
    0.210 & 3332 &  11.46 &  6.1 \\
    0.290 & 3024 &	 11.46 &	5.5  \\
    0.430 & 2649 &  11.94 &  4.8 \\
    0.570 & 2346 &  10.28 &  4.3 \\
    0.710 & 2110 &  11.77 &  3.8 \\
    0.790 & 2032 &  10.16 &  3.7  \\
    0.860 & 1934 &  9.70 &  3.5  \\
    0.930 & 1829 &  9.70 &  3.3 \\
    0.980 & 1284 &  9.23 &  2.3  \\
    1.000 & 872 &  10.87 &  1.6  \\
    \bottomrule
\end{tabular}
\end{table}

\subsection{An Analysis of Out-Of-Distribution Generalization}
\label{section:OOD_generalization}
To validate the out-of-distribution (OOD) generalization capability of our method, we select several pairs of datasets with consistent labels but differing distributions to create the source and target datasets. We train DE$^3$-BERT on the source dataset and subsequently test it on the target dataset.
Specifically, we select the IMDb and SNLI datasets as the target datasets, with the SST-2 and MNLI datasets serving as their corresponding source datasets. The IMDb and SST-2 datasets are designed for sentiment classification tasks, consisting of movie reviews labeled as \textit{positive} or \textit{negative}. Both the SNLI and MNLI datasets are designed for natural language inference tasks, containing sentence pairs labeled as \textit{entailment}, \textit{contradiction}, or \textit{neutral} based on their logical relationships.

Table~\ref{tab:OOD_generalization} presents the experimental results of different early exiting methods on the target datasets. We include the performance of CeeBERT~\cite{ceebert} for comparison, as it is a state-of-the-art cross-domain early exiting method. CeeBERT employs the label score of the prediction probability distribution as its exiting indicator. It learns optimal thresholds for the target datasets in an online and unsupervised manner, enhancing the efficiency of unsupervised cross-domain inference for early exiting BERT.
The experimental results in Table~\ref{tab:OOD_generalization} indicate that our DE$^3$-BERT exhibits slightly inferior performance compared to the state-of-the-art baseline CeeBERT on OOD data.
We attribute this to DE$^3$-BERT's reliance on prior knowledge derived from the training data, i.e., the class prototype representations. In in-distribution generalization, this prior knowledge facilitates reliable exiting decisions on the test data during inference. However, in OOD generalization scenarios, the distributional differences between the training and test data diminish the benefits of this prior knowledge, and it may even compromise the reliability of exiting decisions during inference, affecting the model's performance-efficiency trade-off.

To remedy this, we perform offline adjustments to the class prototype representations in an unsupervised manner before the inference phase.
Specifically, we apply the K-means algorithm~\cite{K-means} to cluster the unlabeled test samples based on their representations suggested by the prototypical networks. The cluster centers are initialized with the original class prototype representations learned from the training data. The final cluster centers obtained from the K-means algorithm serve as adjusted class prototype representations, which align with the distribution of the test data and can be used for distance measurement during the inference phase. Per Table~\ref{tab:OOD_generalization}, we observe that DE$^3$-BERT with adjusted class prototype representations (denoted as DE$^3$-BERT w/ ACP) outperforms the state-of-the-art baseline CeeBERT, demonstrating strong generalization capability on OOD data.
Nevertheless, in scenarios where test samples arrive one by one, the adjustments of class prototype representations must be conducted online, which deserves further research and exploration.

\begin{table}[t]
\caption{Experimental results of different early exiting methods on the target datasets. The datasets are formatted as (source\_target), meaning the dataset preceding the underscore is the source, while the one following the underscore is the target. The results are averaged over 5 different runs.}
\label{tab:OOD_generalization}
\centering
  \begin{tabular}{lll}
    \toprule
Method & SST-2\_IMDb & MNLI\_SNLI\\
    \midrule
    CeeBERT & 81.0 (2.95$\times$) &	 79.4 (2.63$\times$)\\
    DE$^3$-BERT (ours) & -0.4 (2.96$\times$) &	 -0.6 (2.65$\times$)\\
    DE$^3$-BERT w/ ACP (ours) & +0.3 (2.94$\times$) &	 +0.2 (2.63$\times$)\\
    \bottomrule
\end{tabular}
\end{table}

\subsection{Performance on Other Languages}
\label{section:chinese_cls}
To explore the generalization capability of DE$^3$-BERT across different languages, we conduct experiments on two widely used Chinese text classification datasets: LCQMC~\cite{LCQMC} and THUCNews~\cite{THUCNews}.
The LCQMC dataset is designed for sentences-matching tasks, containing Chinese question pairs labeled as 0 or 1 based on their semantic similarity. The THUCNews dataset is designed for sentence-level classification tasks, containing Chinese news articles labeled as 14 categories based on their themes.
We use the pre-trained BERT-base model (\texttt{bert-base-chinese}) released by Google as the backbone and fine-tune it on the LCQMC and THUCNews datasets.
Table~\ref{tab:chinese_cls} presents the experimental results of different early exiting methods. We observe that DE$^3$-BERT achieves a superior performance-efficiency trade-off compared to the competitive baseline GPFEE, validating the generalization capability of our framework across different languages.

\begin{table}[t]
\caption{Experimental results of different early exiting methods on two Chinese classification tasks. The results are averaged over 5 different runs.}
\label{tab:chinese_cls}
\centering
  \begin{tabular}{lll}
    \toprule
Method & LCQMC & THUCNews \\
    \midrule
    BERT-base & 86.7 (1.00$\times$) &	 96.7 (1.00$\times$)\\
    \midrule
    GPFEE & 84.9 (3.01$\times$) &	 96.1 (3.05$\times$)\\
    DE$^3$-BERT (ours) & 85.7 (2.98$\times$) &	 96.6 (3.02$\times$)\\  
    \bottomrule
\end{tabular}
\end{table}

\subsection{Impact of $\lambda$ and $\alpha$}
\label{section:impact_lamda_arfa}
In this section, we conduct experiments on two representative tasks, i.e., SST-2 and QNLI, to investigate the impact of two parameters: the fusion coefficient $\lambda$ (in Equation~(12) in the main paper) and the regularization coefficient $\alpha$ (in Equation~(6) in the main paper).

\subsubsection{Impact of $\lambda$}
\begin{figure}[!t]
\centering
\subfloat[SST-2]{\includegraphics[width=0.46\linewidth]{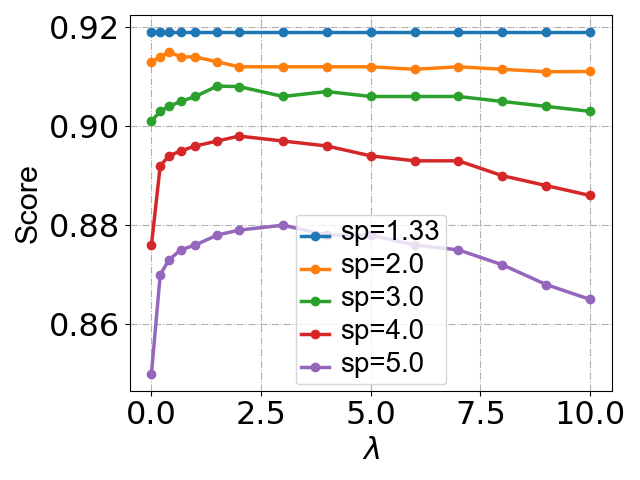}%
\label{fig:lamda_sensitivity_sst2}}
\hfil
\subfloat[QNLI]{\includegraphics[width=0.46\linewidth]{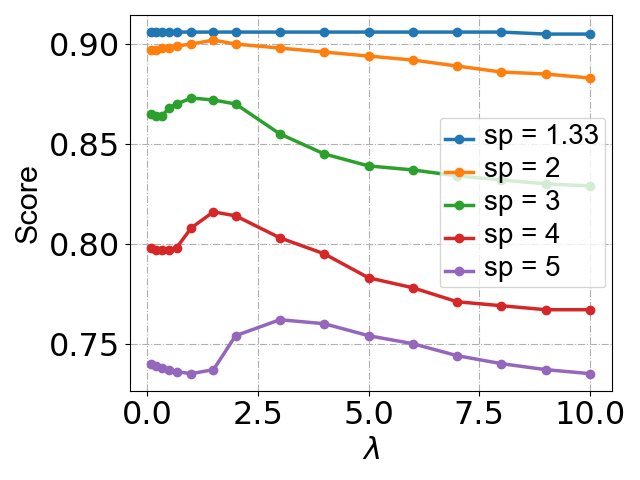}%
\label{fig:lamda_sensitivity_qnli}}
\vspace{-1mm}
\caption{Impact of $\lambda$ on the task performance under different speed-up ratios. Results are on the development sets of SST-2 and QNLI. For each task, we train the model using DAR with a fixed $\alpha$: 0.1 for SST-2, and 0.001 for QNLI. $sp$ denotes the speed-up ratio.}%(a) SST-2. (b) QNLI.
\label{fig:lamda_sensitivity}
% \vspace{-1mm}
\end{figure}

The fusion coefficient $\lambda$ balances the contribution of entropy (local information) and distance ratio (global information) in the exit decision-making process. A higher value of $\lambda$  indicates a stronger emphasis on the distance ratio (global information). We train the models as illustrated in Section~III-B in the main paper with a fixed $\alpha$ value: 0.1 for SST-2, and 0.001 for QNLI, and then evaluate the performance-efficiency trade-off under different values of $\lambda$ during inference. Fig.~\ref{fig:lamda_sensitivity} shows the impact of $\lambda$ on task performance under different speed-up ratios. Note that the experimental results in Fig.~\ref{fig:lamda_sensitivity} include the performance improvements brought by DAR, and we only focus on the impact of $\lambda$ on the model's acceleration performance during inference.

Overall, we observe that, as the value of $\lambda$ increases, the task performance exhibits a consistent pattern of initially increasing and then decreasing under different speed-up ratios. This indicates that the incorporation of the distance ratio can consistently improve the acceleration performance of the model under various speed-up ratios. It also suggests an optimal trade-off between entropy and distance ratio. Additionally, it is noticeable that as the speed-up ratio increases, the optimal value of $\lambda$ tends to increase, and compared to entropy-based exiting strategies ($\lambda =0$), the introduction of distance ratio yields more significant performance improvements. We attribute this to the following reasons. Under high speed-up ratios, samples tend to exit at shallow layers (see Appendix C-A). The limited capability of shallow classifiers amplifies the gap between the confidence level reflected by entropy and the correctness of early predictions. This makes it crucial to introduce the distance ratio (global information) to improve the estimation of prediction correctness, thus enhancing the reliability of exiting decisions. Consequently, the performance gains brought by distance ratio tend to be more significant. This explanation is subsequently demonstrated through experiments, as detailed in Appendix \ref{section:estimation_accuracy}. Finally, for parameter selection, the optimal $\lambda$ value differs across tasks and speed-up ratios, as shown in Fig.~\ref{fig:lamda_sensitivity}. Therefore, it is recommended to conduct a parameter search within the range between 0 and 10 according to specific scenarios.

\subsubsection{Impact of $\alpha$} 
\begin{figure}[!t]
\centering
% \vspace{-3mm}
\subfloat[SST-2]{\includegraphics[width=0.5\linewidth]{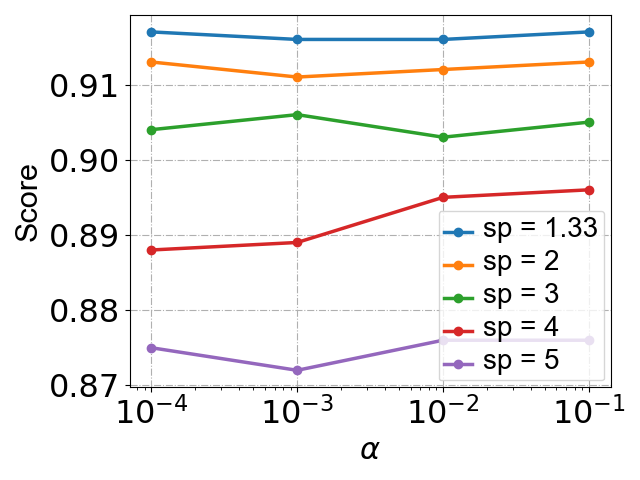}%
\label{fig:arfa_sensitivity_sst2}}
\hfil
\subfloat[QNLI]{\includegraphics[width=0.5\linewidth]{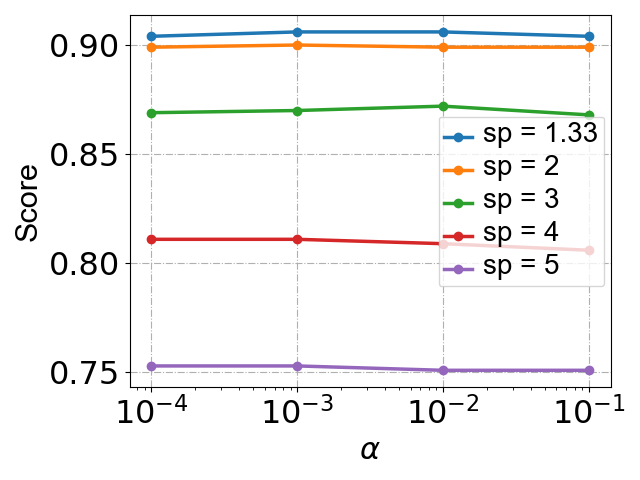}%
\label{fig:arfa_sensitivity_qnli}}
\vspace{-1mm}
\caption{Impact of $\alpha$ on the task performance under different speed-up ratios. Results are on the development sets of SST-2 and QNLI. For each task, we employ the hybrid exiting strategy during inference with a fixed $\lambda$: 1.0 for SST-2, and 2.0 for QNLI. $sp$ denotes the speed-up ratio.}% (a) SST-2. (b) QNLI.
\label{fig:arfa_sensitivity}
% \vspace{-2mm}
\end{figure}

The regularization coefficient $\alpha$ balances the cross-entropy loss and DAR during training. A higher value of $\alpha$ indicates a stronger emphasis on optimizing DAR. We train the models as illustrated in Section~III-B in the main paper, with $\alpha$ values of \{0.0001, 0.001, 0.01, 0.1\}. During inference, the proposed hybrid exiting strategy is employed with a fixed $\lambda$ value of 1.0 for SST-2 and 2.0 for QNLI. Fig.~\ref{fig:arfa_sensitivity} shows the impact of $\alpha$ on task performance under different speed-up ratios. It can be observed that the optimal value of $\alpha$ differs across tasks and speed-up ratios, but the performance changes caused by different $\alpha$ values are always less than 1$\%$, indicating that the acceleration performance is not significantly affected by $\alpha$ selection. Moreover, an $\alpha$ value between 0.0001 and 0.1 can always lead to satisfactory performance on different tasks.

\begin{figure}[!t]
\centering
% \vspace{-3mm}
\subfloat[SST-2]{\includegraphics[width=0.45\linewidth]{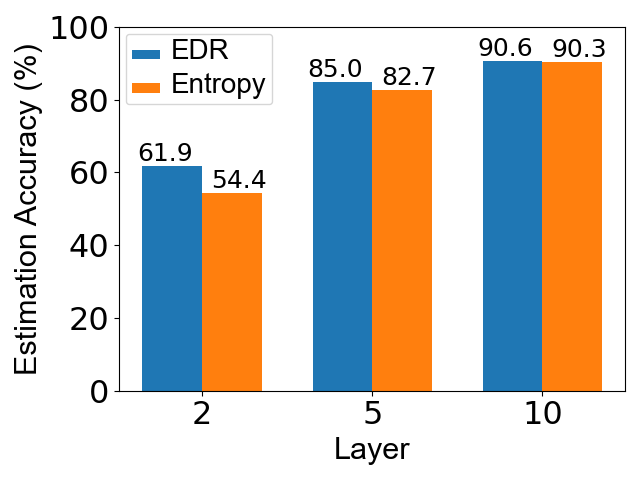}%
\label{fig:correctness_estimation_sst2}}
\hfil
\subfloat[QNLI]{\includegraphics[width=0.45\linewidth]{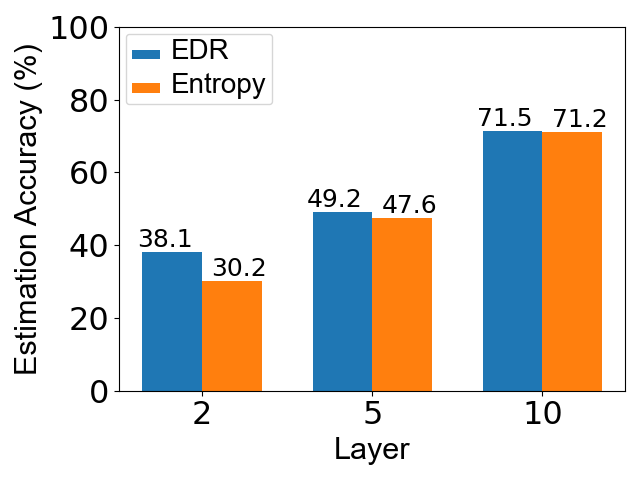}%
\label{fig:correctness_estimation_qnli_0.2}}
\hfil
\vspace{-3mm}
\subfloat[QQP]{\includegraphics[width=0.45\linewidth]{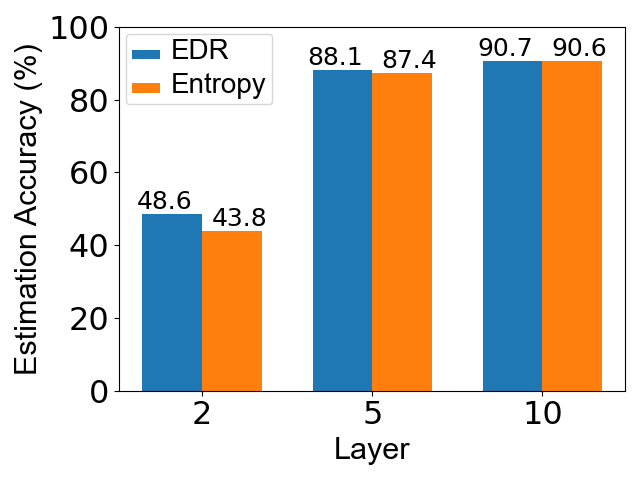}%
\label{fig:correctness_estimation_qqp_0.2}}
\hfil
\subfloat[MNLI]{\includegraphics[width=0.45\linewidth]{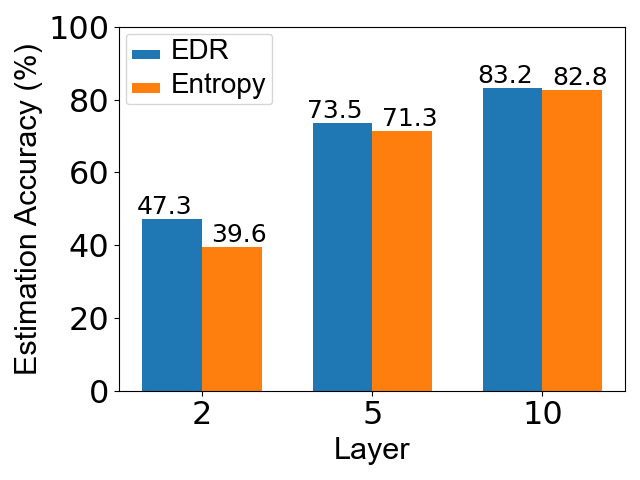}%
\label{fig:correctness_estimation_mnli_0.2}}
\caption{Accuracy of prediction correctness estimation based on entropy and our hybrid exiting indicator (the $\mathrm{EDR}$ value) on layers 2, 5, and 10. Results are on the development sets of four GLUE tasks.}% (a) SST-2. (b) QNLI. (c) QQP. (d) MNLI.
\label{fig:correctness_estimation}
% \vspace{-2mm}
\end{figure}

\subsection{Accuracy of Prediction Correctness Estimation} 
\label{section:estimation_accuracy}
To further investigate the effect of distance ratio (global information) on the estimation of prediction correctness, we quantitatively analyze the estimation accuracy based on entropy and our hybrid exiting indicator ($\mathrm{EDR}$ value), respectively. Specifically, we first estimate the prediction correctness by comparing the entropy or $\mathrm{EDR}$ value with the predefined threshold $\tau$. An entropy or $\mathrm{EDR}$ value below (above) the threshold indicates a correct (incorrect) early prediction. Then, we compare the early prediction with the sample's ground-truth label to obtain the true value of prediction correctness. On this basis, we finally derive the accuracy of prediction correctness estimation. Fig.~\ref{fig:correctness_estimation} shows the experimental results at the 2nd, 5th, and 10th layers on a representative subset of GLUE, where the threshold $\tau$ is set to 0.2.

Overall, we observe that the $\mathrm{EDR}$ value consistently outperforms the entropy in estimating the correctness of early predictions across different layers and tasks. This demonstrates that distance-based global information can complement entropy-based local information, offering a more accurate estimation of prediction correctness. Additionally, it is noticeable that as the number of layers decreases, the accuracy of prediction correctness estimation based on both entropy and $\mathrm{EDR}$ values tends to decline, and the introduction of distance ratio yields more significant improvements in estimating prediction correctness. We believe that this observation can be attributed to the following reasons. The limited representation capability of internal classifiers in shallow layers leads to insufficient local information based on entropy, exacerbating the difficulty of estimating prediction correctness. Consequently, compared to deeper layers, introducing distance ratio at shallow layers is particularly crucial for enhancing the accuracy of prediction correctness estimation. This, in turn, contributes to more reliable exiting decisions and better trade-offs between model performance and efficiency. This further explains our observations in Fig.~4 in the main paper and Fig.~\ref{fig:lamda_sensitivity}, that as the speed-up ratio increases, the introduction of distance ratio tends to yield greater performance improvements.

% in task performance

Based on the above analysis, we can conclude that the introduction of the distance ratio can effectively improve the estimation of prediction correctness (especially for shallow layers), which further enhances the reliability of exiting decisions, leading to a better trade-off between model performance and inference efficiency.

\subsection{Storage and Computational Costs}
\label{section:storage_compute_overheads}

\subsubsection{Parameter Volumes and Training Time Costs} 
Table~\ref{tab:storage_training_costs} compares the parameter volumes and training time costs of different models.
Building on the classic entropy-based early exiting model DeeBERT~\cite{Deebert}, which adds only an internal classifier at each intermediate layer, our DE$^3$-BERT additionally incorporates an prototypical network at each intermediate layer for distance calculation. Despite this enhancement, compared to the original BERT-base, DE$^3$-BERT incurs only a $5.95\%$ increase in parameter volumes for $K=2$ and $5.96\%$ for $K=3$, demonstrating its storage efficiency. Notably, the number of additional parameters is not significantly affected by the number of classes $K$, as such storage overhead is primarily driven by the prototypical networks, which are $K$-agnostic.
Furthermore, despite additional parameters, training DE$^3$-BERT is not significantly slower than training DeeBERT or BERT-base, and it is even faster on some tasks such as MRPC and QQP. This can be attributed to the additional supervision provided by intermediate-layer losses, i.e., the cross-entropy loss for internal classifiers and the DAR for prototypical networks, which accelerates the model convergence during training.
Therefore, our method can significantly reduce inference costs, with minimal additional parameters and comparable training time costs compared to DeeBERT and BERT-base.

\begin{table}[!t]
\caption{Comparison of parameter volumes and training time costs. Training time refers to the cost incurred from the start of training until the best checkpoint on the development set is achieved and is collected under identical conditions. $K$ denotes the number of classes.}
\label{tab:storage_training_costs}
\centering
\setlength{\tabcolsep}{1mm}
\begin{tabular}{lccccccccc}
    \toprule
    \multirow{2}{1.0cm}{Model} & & \multicolumn{2}{c}{$\#$Params}  & \multicolumn{5}{c}{Training Time (min)}   \\
    & & $K=2$ & $K=3$ & & MRPC & SST-2 & QQP & MNLI \\
    \midrule
    BERT-base &  &	109.48M       & 109.48M  &  & 1.4 &	38  &	313 &	166     \\
    DeeBERT &  &	$+$16.92K       & $+$25.38K  &  & 1.3 &	34  &	304 &	163     \\
    DE$^3$-BERT (ours) &  &	$+$6.51M      & $+$6.52M	&  & 0.95 & 44 &	293 &	165     \\

    \bottomrule
\end{tabular}
\vspace{-2mm}
\end{table}

\subsubsection{Computational Complexity} 
Table~\ref{tab:flops_analysis} shows each module's computational complexity within the  DE$^3$-BERT model.
The computational complexity incurred by making exiting decisions mainly arises from executing internal classifiers and prototypical networks, as well as $\rm EDR$ calculations. This overhead is approximately 1.21M per layer, which is negligible compared to a single encoder block (1813.5M), resulting in only a $0.067\%$ increase in computational costs. Consequently, our DE$^3$-BERT framework incurs negligible computational overhead during the exit decision-making process.
Moreover, it can be observed that the computational overhead incurred by making exiting decisions remains almost unchanged with respect to the number of classes $K$. This can be attributed to the following factors: (a) The prototypical network is $K$-agnostic, since it projects the hidden states of each layer to a metric space; (b) The computation of EDR is also independent of $K$, as our method only measures the distances between the sample representation and the prototypes of the top-2 predicted classes, instead of all class prototypes; (c) Although the cost of the internal classifier scales linearly with $K$, it is negligible compared to that of the prototypical network. These observations confirm that the proposed method introduces minimal additional computation costs and maintains high efficiency even as the number of classes increases.
Furthermore, as shown in Table~\ref{tab:flops_analysis}, the encoder block dominates the network's computational overhead, making the total inference costs approximately proportional to the number of executed layers. This confirms the validity of using saved layers to evaluate model acceleration.

\begin{table}[!t]
\caption{Computational complexity of each module in the DE$^3$-BERT model. $K$ denotes the number of classes, and the prediction head denotes the classifier connected to the final layer.}
\label{tab:flops_analysis}
\centering
\begin{tabular}{lcc}
\toprule
\multirow{2}{3.0cm}{Module} & \multicolumn{2}{c}{FLOPs} \\
 & $K=2$ & $K=3$ \\
\midrule
Embedding & 786.4K & 786.4K\\
Encoder & 1813.5M & 1813.5M\\
Pooler & 1.2M & 1.2M\\
Prediction Head & 3.1K & 4.6K\\
Prototypical Network & 1.2M & 1.2M\\
Internal Classifier & 3.1K & 4.6K\\
$\rm EDR$ Calculation & 9.2K & 9.2K\\
\bottomrule
\end{tabular}
\end{table}

\subsection{Visualization of Sample Representations}
\label{section:visualization}
\begin{figure}[!t]
\centering
\vspace{-5mm}
\subfloat[]{\includegraphics[width=0.45\linewidth]{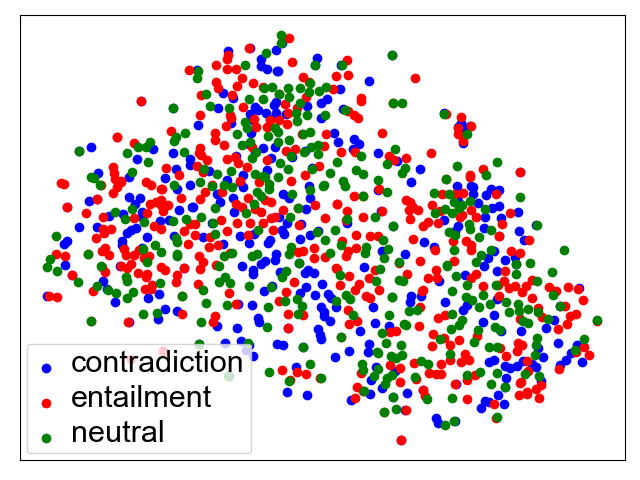}%
\label{fig:visualization_2_wo}}
\hfil
\subfloat[]{\includegraphics[width=0.45\linewidth]{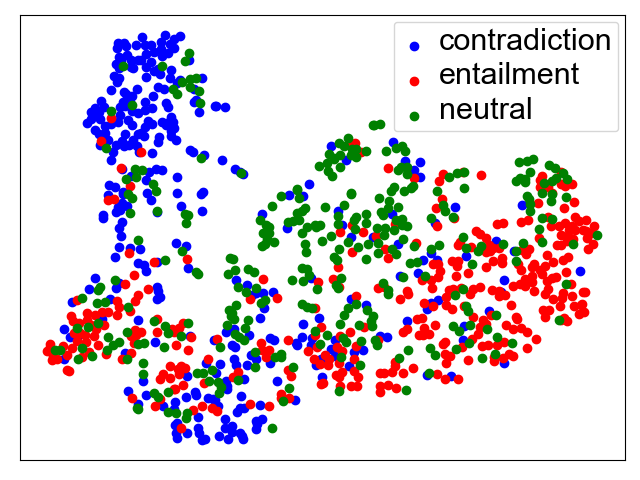}%
\label{fig:visualization_2_w}}
\hfil
\vspace{-3mm}
\subfloat[]{\includegraphics[width=0.45\linewidth]{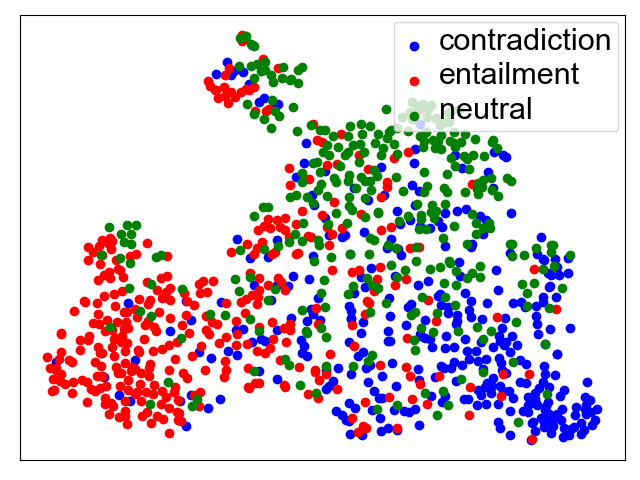}%
\label{fig:visualization_5_wo}}
\hfil
\subfloat[]{\includegraphics[width=0.45\linewidth]{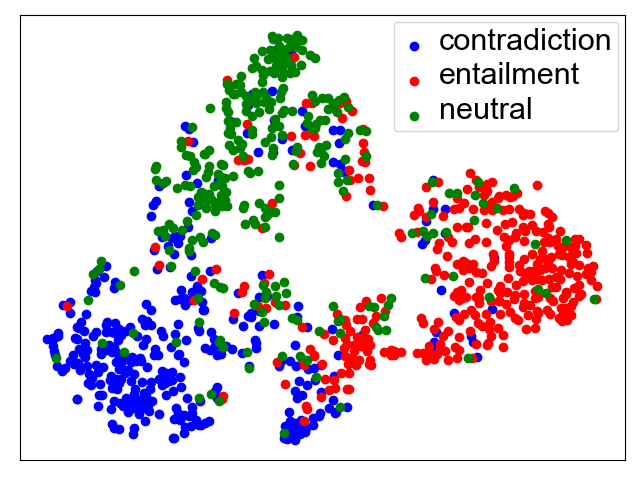}%
\label{fig:visualization_5_w}}
\caption{t-SNE visualization of the sample representations generated by the models trained with and without DAR on layers 2 and 5. Our DE$^3$-BERT model trained with DAR produces more discriminative sample representations, further strengthening the effect of DAR in generating high-quality metric spaces. (a) DE$^3$-BERT w/o DAR on layer 2. (b) DE$^3$-BERT on layer 2. (c) DE$^3$-BERT w/o DAR on layer 5. (d) DE$^3$-BERT on layer 5.}
\label{fig:visualization}
\vspace{-2mm}
\end{figure}

To verify whether DAR facilitates the learning of metric space and class prototype representations, we use the t-SNE projection~\cite{tsne} to visualize the sample representations generated by the prototypical networks of the DE$^3$-BERT model. For comparison, we train a variant (denoted by DE$^3$-BERT w/o DAR) of the DE$^3$-BERT model using only cross-entropy loss and visualize the sample representations obtained by pooling the hidden states of encoder layers (i.e., inputs to internal classifiers). Fig.~\ref{fig:visualization} shows the visualization results on layers 2 and 5 of the two models on the MNLI development set. We can observe that, on both layers 2 and 5, DAR encourages clearer classification boundaries between different classes. This indicates that DAR can facilitate the learning of high-quality metric space and class prototype representations, which is crucial for accurately estimating the prediction correctness and making reliable exiting decisions, as illustrated in Fig.~5 in the main paper.

\subsection{Additional Effect of Prototypical Networks}
\label{section:PN_homo}
As mentioned in Section~III-A in the main paper, the main purpose of introducing the prototypical networks is to learn the class prototype representations and the associated metric space. Additionally, we believe that the introduction of prototypical networks can help avoid the homogeneity issue between entropy and distance ratio since they are computed in different representation spaces. This enables them to mutually correct each other. To verify our claim, we analyze the Spearman correlation coefficient between the entropy and distance ratio of our DE$^3$-BERT models with and without prototypical networks, respectively. Concretely, we train a variant (denoted by DE$^3$-BERT w/o PN) of our DE$^3$-BERT model as illustrated in Section~III-B in the main paper. The computation of DAR terms and the updating of class prototype representations during training, as well as the calculation of distance ratios during inference, are all based on the sample representations obtained by pooling the hidden states of encoder layers (i.e., inputs to internal classifiers). We then analyze the Spearman correlation coefficient between the entropy and distance ratio of this variant and the full DE$^3$-BERT model. Table~\ref{tab:PN_homo} illustrates the analysis results on a representative subset of GLUE based on the model outputs at the second layer. The results indicate a significant increase in correlation between entropy and distance ratio when the prototypical networks are removed. These results are consistent with the results shown in Fig.~5 in the main paper that removing the prototypical networks impairs the acceleration performance of the model, which further confirms that the prototypical networks can effectively avoid the homogeneity between entropy and distance ratio, thus enhancing their integration.

\begin{table}[!t]
\caption{The impact of prototypical networks (PN) on the correlation between entropy and distance ratio. We report the Spearman correlation coefficient between entropy and distance ratio based on the model outputs at layer 2 on the development sets of four GLUE tasks.}%. PN denotes the prototypical networks.
\label{tab:PN_homo}
\centering
  \begin{tabular}{lcccc}
    \toprule
    Model & QQP &  MNLI  & SST-2 &  QNLI  \\
    \midrule
    DE$^3$-BERT & 0.207 &	 0.018 &	0.145 &	 0.066  \\
    DE$^3$-BERT w/o PN & 0.569 &  0.389 &  0.479 & 0.422 \\
    \bottomrule
\end{tabular}
% \vspace{-2mm}
\end{table}

\begin{table}[!t]
\caption{Range for each hyperparameter and the corresponding values selected for all tasks. H denotes the hyperparameter, $bsz$ denotes the batch size, and $lr$ denotes the learning rate.}
\label{tab:hyperparameter_selection}
\centering
\scalebox{0.83}{
  \begin{tabular}{cccccccc}
    \toprule
    H & Range &  MNLI & MRPC  & QNLI & QQP & RTE & SST-2  \\
    \midrule
    $bsz$ & 16, 32, 128 &	32  & 32 & 32 & 32 & 16 & 32  \\
    $lr$ & 1e-5, 2e-5, 3e-5, 5e-5 &	2e-5  & 3e-5 & 3e-5 & 2e-5 & 2e-5 & 2e-5  \\
    $\alpha$ & 1e-4, 1e-3, 1e-2, 1e-1 &	1e-1  & 1e-4 & 1e-3 & 1e-2 & 1e-1 & 1e-1  \\
    $\lambda$ & 0.667, 1.0, 1.5, 2.0, 3.0 &	0.667  & 1.5 & 2.0 & 0.667 & 1.5 & 1.0  \\

    \bottomrule
\end{tabular}
}
\end{table}

\subsection{Experimental Results of Different DAR Forms}
\label{section:DAR_comparison}
Table~\ref{tab:dar_performance} shows the performance comparison of different DAR forms on the GLUE test sets under the speed-up ratios of $2.0\times$ and $3.0\times$. Overall, we observe that different forms of DAR yield comparable results, all surpassing the competitive baseline GPFEE~\cite{globalpast}. We further demonstrate the statistical significance of these improvements through one-tailed t-tests. These findings confirm the effectiveness and robustness of our framework across various metric learning methods.
Specifically, as the speed-up ratio increases, Center Loss exhibits a slight advantage over the other two DAR forms in terms of performance degradation, yielding a better trade-off between model performance and efficiency, although the performance gaps are not substantial. We speculate that this is caused by the conflicting optimization of Alienation Loss and cross-entropy loss during the training phase. Concretely, both Alienation Loss and cross-entropy loss have the effect of enlarging the difference between intra-class and inter-class distances, but their different implementation mechanisms lead to conflicting optimization during training. As a result, this affects the learning of metric space and class prototype representations, which ultimately compromises the acceleration performance of the model. In contrast, Center Loss, which aims to reduce intra-class distances, is better suited to cross-entropy loss. This facilitates the learning of high-quality metric space and class prototype representations, thus improving the acceleration performance of the model. Therefore, considering the above analysis and the computational efficiency, we primarily use Center Loss in DAR.% in this paper.

\begin{table*}[t]
\caption{Model performance on the GLUE test sets with different DAR forms. The speed-up ratios are 2.0$\times$ and 3.0$\times$. DE$^3$-BERT, DE$^3$-BERT-v2, and DE$^3$-BERT-v3 use the Center Loss, Alienation Loss, and Combined Loss in DAR, respectively. We include the competitive baseline GPFEE~\cite{globalpast} in comparison. All baseline results are taken from the original paper. Best performance values are marked in bold. The * denotes a statistically significant performance improvement over the competitive baseline GPFEE at a significance level of 0.05.}
\label{tab:dar_performance}
\centering
  \begin{tabular}{clccccccl}
    \toprule
    & \multirow{2}{1.0cm}{Method}   &  MNLI  & MRPC  &  QNLI  &  QQP  &  RTE  &  SST-2  &  \multirow{2}{0.5cm}{AVG}\\
    & &  Acc & F1/Acc  & Acc  & F1/Acc   & Acc   & Acc   &                          \\

    \midrule
    & BERT-base & 84.6 (1.00$\times$) & 88.9/- (1.00$\times$)& 90.5 (1.00$\times$) & 71.2/- (1.00$\times$)& 66.4 (1.00$\times$) & 93.5 (1.00$\times$) &-\\
    \midrule
    \multirow{4}{0.1cm}{\rotatebox{90}{$\sim$2$\times$}} 
    & GPFEE & 83.3 (1.96$\times$) & 
    {\bf 87.0/81.8} (1.98$\times$) & 
    89.8 (1.97$\times$) & 
    71.2/89.4 (2.18$\times$) &
    64.5 (2.04$\times$) & 
    {\bf 92.8} (2.02$\times$)&
    82.5 \\  
    & DE$^3$-BERT (ours) &	83.2 (2.04$\times$)& 	
    86.6/81.5 (1.98$\times$)&	
    90.0 (2.07$\times$)& 	
    71.2/89.4 (2.16$\times$)&	
    65.7 (1.99$\times$)&	
    92.5 (2.02$\times$)&
    {\bf 82.6}$^*$ \\
    & DE$^3$-BERT-v2 (ours) &	{\bf 83.4} (2.02$\times$)&	86.5/81.2 (1.97$\times$)&	{\bf{90.1}} (2.04$\times$)&	71.0/89.4 (2.12$\times$)&	{\bf{65.9}} (2.03$\times$)&	92.4 (2.01$\times$)&  {\bf 82.6}$^*$\\
    & DE$^3$-BERT-v3 (ours) &	83.1 (2.05$\times$)&	86.5/81.3 (1.99$\times$)&	{\bf 90.1} (1.98$\times$)&	{\bf{71.4/89.4}} (2.17$\times$)&	65.8 (2.05$\times$)&	92.5 (2.04$\times$)& {\bf 82.6}$^*$\\
    \midrule
    \multirow{4}{0.1cm}{\rotatebox{90}{$\sim$3$\times$}} 
    & GPFEE & 78.4 (2.99$\times$) & 
    {\bf 84.5/77.7} (2.87$\times$) & 
    {\bf 87.3} (2.78$\times$) & 
    70.4/89.2 (3.16$\times$) &
    63.0 (2.88$\times$) & 
    91.1 (2.97$\times$)& 80.1\\     
    & DE$^3$-BERT (ours) &	79.9 (2.99$\times$)& 	
    83.8/76.5 (3.01$\times$)&	
    87.0 (3.02$\times$)& 	
    {\bf 70.6/89.3} (3.18$\times$)&	
    {\bf 63.6} (2.97$\times$)&	
    {\bf 91.4} (2.98$\times$)&  {\bf 80.3}$^*$\\
    & DE$^3$-BERT-v2 (ours) &	{\bf{80.1}} (3.01$\times$)&	84.0/76.7 (3.02$\times$)&	86.7 (2.98$\times$)&	70.4/89.2 (3.19$\times$)&	63.4 (2.99$\times$)&	91.1 (2.98$\times$)& 80.2$^*$\\
    & DE$^3$-BERT-v3 (ours) &	80.0 (3.02$\times$)&	83.8/76.3 (2.99$\times$)&	86.8 (3.01$\times$)&	70.5/89.3 (3.18$\times$)&	63.5 (2.98$\times$)&	91.2 (3.01$\times$)& 80.2$^*$\\
   
    \bottomrule
\end{tabular}
\vspace{-1mm}
\end{table*}

\section{Hyperparameter Settings}
\label{section:hyperparameter_selection}
To enhance the reproducibility of our results, we present the range for each hyperparameter along with the corresponding values selected for all tasks, as shown in Table~\ref{tab:hyperparameter_selection}.

\section{Discussion}
\label{section:discussion}
\subsection{Discussion of Experimental Results}
In our study, we investigate the impact of integrating distance-based global information into the exit decision-making process. Accordingly, we introduce the distance ratio (global information) into the classic entropy-based early exiting method. Our experimental results indicate that incorporating the distance ratio can effectively improve the estimation of prediction correctness. This further facilitates more reliable exiting decisions, yielding a better trade-off between model performance and efficiency. Specifically, the performance improvements attributed to the distance ratio appear to be more significant in high acceleration scenarios.

In general, the experimental results demonstrate our hypothesis that the performance bottleneck of existing early exiting methods primarily stems from their exclusive reliance on local information from individual test samples to make exiting decisions, while overlooking the global information provided by the sample population. We believe that the latent class information of a test sample can be reflected by its distances from class prototypes, which can effectively complement the entropy-based local information and improve the accuracy of prediction correctness estimation from a global perspective. In this way, the proposed DE$^3$-BERT approximates the behavior of the oracle, thus achieving a better trade-off between model performance and efficiency. Furthermore, the limited representation capability of shallow internal classifiers leads to insufficient local information based on entropy, which emphasizes the importance of introducing global information. This explains the superiority of our hybrid exiting strategy in high acceleration scenarios. 

Our study provides a new perspective on estimating the prediction correctness for early exiting models, which inspires facilitating reliable exiting decisions by integrating distance-based global information.

\subsection{Limitations and Future Work}
\label{section:limitations}
For further research in the near future, we comprehensively discuss the limitations of this work from three perspectives. Firstly, our DE$^3$-BERT framework caters to classification tasks only since DE$^3$-BERT incorporates the distance ratio that requires the prototype representations of all classes, which are unavailable for other non-classification tasks. One of our future studies is to extend our DE$^3$-BERT framework to more sophisticated regression tasks by training a neural network to map continuous regression values to their corresponding prototype representations. Secondly, our DE$^3$-BERT framework improves the entropy-based early exiting by transferring knowledge from the training set to the test set via class prototypes, assuming that the distribution is identical between the training and test data, i.e., in-distribution generalization. The out-of-distribution case brings extreme challenges to DE$^3$-BERT, which is worth further research and exploration. Finally, different from our DE$^3$-BERT which focuses on improving the exiting strategies, some research, such as CascadeBERT~\cite{li2021cascadebert}, GPFEE~\cite{globalpast}, LeeBERT~\cite{zhu2021leebert}, and COSEE~\cite{COSEE}, primarily address different issues of early exiting, including training schemes, model architecture, and model calibration. Exploring the combination of our DE$^3$-BERT with the aforementioned methods would be an interesting endeavor, as it contributes to further enhancing the acceleration performance of early exiting models by investigating the potential orthogonality and complementarity between different techniques.

\subsection{Real-World Applications and Implications}
\label{section:real-world}
\subsubsection{Real-World Applications}
Our method effectively accelerates the inference of PLMs by incorporating distance-based global information. It is well-suited for low-latency applications, including chatbots, recommendation systems, and real-time text analysis. For instance, in customer service scenarios, our method can effectively enhance user satisfaction by facilitating immediate interaction with customers.

\subsubsection{Real-World Implications}
The proposed method enhances the deployment of PLMs in resource-constrained devices like mobile phones by conserving energy and reducing operational costs, broadening access to advanced AI technologies. It also promotes more sustainable and eco-friendly practices in artificial intelligence. Overall, this method enables faster, broader, and more sustainable AI solutions.

\subsection{Scalability of the DE$^3$-BERT Framework}
\label{section:scalability}
\subsubsection{Scalability on Larger Datasets}
Concerning the scalability of DE$^3$-BERT on larger datasets, it is crucial to analyze its computational overhead in both the training and inference phases. Firstly, our DE$^3$-BERT introduces negligible training overhead compared to the original BERT-base. Specifically, at each training step, the computational complexity of BERT is $O(B(n^2 d + n d^2))$ per layer, where $B$ is the batch size, $n$ is the sequence length, and $d$ is the embedding dimension.
In comparison, the computational overhead introduced by DE$^3$-BERT arises from the incorporation of prototypical networks and internal classifiers, along with the calculation of intermediate-layer losses and updates to class prototype representations, resulting in a complexity of $O(Bd^2)$ per layer.
Hence, the computational complexity of DE$^3$-BERT is on par with that of BERT, both being $O(B(n^2 d + n d^2))$ per layer for each training step.
Additionally, the results in Table~\ref{tab:storage_training_costs} confirm that incorporating intermediate-layer losses may accelerate model convergence, i.e., training DE$^3$-BERT is not significantly slower than training the original BERT-base and might even be faster. Secondly, DE$^3$-BERT significantly reduces the model's inference costs by enhancing the reliability of exiting decisions while introducing negligible computational overhead during the exit decision-making process, as shown in Table~\ref{tab:flops_analysis}.

Based on the above analysis, our DE$^3$-BERT is computationally efficient in both training and inference, ensuring its potential for scaling to larger datasets.

\subsubsection{Scalability on Regression Tasks}
Classification and regression are two fundamental tasks. As discussed in Appendix~\ref{section:limitations}, DE$^3$-BERT caters to classification tasks and may not be directly applied to more complicated regression tasks due to the lack of class prototypes. Nevertheless, with appropriate adjustments to the distance metric, our framework can be applicable to regression tasks.
Specifically, a preliminary solution involves extending the concept of prototype representations from discrete classes to continuous regression values by using a neural network to map each regression value to its corresponding prototype representation. During inference, we can estimate prediction correctness from a global perspective by comparing the distances between the sample representation and the prototype representations of various regression values, including the predicted value. This facilitates informed exiting decisions and warrants further research and exploration.

\end{document}